%% file: main.tex
\documentclass[11pt,a4paper]{article}
\input{preamble}
\begin{document}
\input{manuscript}
\clearpage
\input{supplement}
\end{document}

%% file: preamble.tex
\usepackage[T1]{fontenc}
\usepackage[utf8]{inputenc}
\usepackage{lmodern}
\usepackage[margin=24mm]{geometry}
\usepackage{graphicx}
\usepackage{pdflscape}
\usepackage{needspace}
\usepackage{booktabs,tabularx,longtable,array}
\usepackage{amsmath,amssymb}
\usepackage{enumitem}
\usepackage{caption}
\usepackage{microtype}
\usepackage{fancyhdr}
\usepackage[hidelinks]{hyperref}
\hypersetup{pdftitle={Aegis: A Conversational Prototype for Lung Cancer Risk Information and Follow-up},pdfauthor={Han Bowen, Cao Chengwei, Zhang Runxin, Jiang Yuyang}}
\setlist{nosep,leftmargin=*}
\newcolumntype{Y}{>{\raggedright\arraybackslash}X}

%% file: manuscript.tex
\title{\vspace{-1.3em}Aegis: A Conversational Prototype for Lung Cancer\newline Risk Information and Follow-up}
\author{Han Bowen\textsuperscript{\textdagger,*}\quad Cao Chengwei\textsuperscript{\textdagger}\quad Zhang Runxin\textsuperscript{\textdagger}\quad Jiang Yuyang\textsuperscript{\textdagger}\\[5pt]
\small School of Computer Sciences, Universiti Sains Malaysia\\
\small 11800 USM, Penang, Malaysia\\[4pt]
\small \textsuperscript{\textdagger}All authors contributed equally and share first authorship.\\
\small \textsuperscript{*}Corresponding author: \href{mailto:hanbowen@student.usm.my}{hanbowen@student.usm.my}}
\date{}
\maketitle
\vspace{-1.5em}
\begin{abstract}
Collecting self-reported risk information and maintaining continuity between assessment, explanation, and follow-up are practical components of a lung cancer screening workflow. We present Aegis, a browser-based research prototype that combines a conversational questionnaire, rule-based risk categorization, downloadable reports, retrieval-augmented health information, and doctor--user communication. A session-oriented controller manages question progression and clarification, while a language model assists with answer validation. Structured responses are passed to a separate risk-analysis module, and reports are linked to user and session records through Firebase services. The project adopts a multi-agent design that separates information collection, validation, assessment, and report generation. Its documented verification scope comprises 45 functional test specifications covering account management, questionnaire interaction, reporting, doctor-side review, and consultation. The available evaluation describes workflow behavior qualitatively; it provides no per-case execution results, clinical cohort, predictive-performance measurements, or comparative experiments. We therefore present Aegis as an engineering prototype for organizing risk information and follow-up, with clinical validity and the benefits of its agent-based design remaining unestablished. The paper describes the implementation, distinguishes implemented workflows from incomplete extensions, and identifies the evidence required for reproducible and clinically meaningful evaluation.
\end{abstract}

\section{Introduction}
Lung cancer accounted for the largest share of cancer deaths worldwide in the 2022 estimates reported by Bray et al.~\cite{bray}. The National Lung Screening Trial established a mortality benefit for low-dose computed tomography (LDCT) relative to chest radiography in its high-risk study population~\cite{nlst}. Implementing LDCT screening requires healthcare infrastructure and economic resources, creating particular challenges in resource-constrained settings~\cite{saul}. A digital questionnaire can help organize information before a clinical encounter, but its availability alone does not establish an effective cancer-screening intervention.

Aegis addresses the software workflow around self-reported information: collecting answers, requesting clarification, presenting a preliminary risk category, preserving reports, and connecting users with further information or a doctor-facing interface. The intended users include adults managing their own health, family members assisting with registration, and doctors reviewing records. Text and voice interaction are included to support different input preferences. Their effects on accessibility and completion rates have not been measured.

The engineering challenge is to maintain continuity across these steps without conflating their outputs. A semantically relevant answer can still be medically inaccurate; a rule-generated category can be internally consistent without being clinically calibrated; and a generated explanation can be fluent without being correct. Aegis separates conversational data collection, rule-based assessment, information retrieval, and human consultation at the module level. This paper describes that organization and the available verification evidence. Its contribution is an integrated prototype and a traceable account of the workflow, rather than a new validated risk model or a demonstrated improvement in screening outcomes.

\section{Related Work and Scope}
\subsection{Screening, risk estimation, and imaging}
The National Lung Screening Trial and NELSON trial provide randomized evidence concerning LDCT screening in their respective study populations~\cite{nlst,nelson}. Screening recommendations translate such evidence into eligibility and follow-up guidance; the 2021 US Preventive Services Task Force statement is one example~\cite{uspstf}. A recent Malaysian surgical consensus also addresses screening alongside diagnosis, staging, treatment, and surveillance~\cite{malaysia}. These clinical pathways establish the context for information collection, but they do not validate a particular digital questionnaire. Infrastructure and economic constraints remain relevant when considering how screening services can be implemented~\cite{saul}.

Questionnaire-based risk estimation has a separate methodological basis. Tammem\"agi et al. developed and validated the PLCO\textsubscript{M2012} model for selecting individuals for lung cancer screening~\cite{tammemagi}. Aegis does not document an implementation of that model. Its local rules produce low, medium, or high categories without a specified prediction horizon, reported calibration, or comparison against an established model. We consequently treat these categories as prototype outputs, not estimated probabilities of cancer or validated screening-eligibility decisions. The distinction matters when a user sees a category repeatedly: consistency of display does not establish validity of the underlying estimate.

Imaging-based approaches address another task. Ardila et al. studied end-to-end lung cancer screening with three-dimensional deep learning on LDCT~\cite{ardila}, while Mikhael et al. introduced Sybil for predicting future lung cancer risk from a single LDCT examination~\cite{sybil}. Those studies concern models operating on imaging data. Aegis principally collects self-reported answers and organizes reports and communication. Its CT-related screens are retained in the supplement as design artifacts; they do not provide evidence of a comparable imaging model or its validation.

\subsection{Medical question answering and conversational systems}
Medical language-model research spans tasks with different inputs and evaluation criteria. MedQA supplies questions from medical examinations~\cite{medqa}; Singhal et al. investigated clinical knowledge in large language models~\cite{singhal}. More recent work by Tu et al. evaluates conversational diagnostic AI in simulated consultations~\cite{amie}, and Li\'evin et al. study conversational AI for disease management~\cite{management}. Together, these lines of work distinguish answering a self-contained question from conducting a dialogue in which information must be elicited and used across turns.

Aegis combines an interview-like questionnaire with a separate explanatory assistant. Its assessment dialogue is directed toward collecting fields for a rule-based module, rather than generating an unrestricted diagnosis. The relevant engineering questions therefore include whether the next question is appropriate, whether a clarification preserves prior answers, and whether the final record matches the user's responses. Performance on medical examination questions would not answer these workflow questions. Conversely, completing a questionnaire would not demonstrate the medical reasoning quality evaluated in diagnostic-dialogue research.

\subsection{Agent organization and task execution}
MetaGPT organizes language-model agents through differentiated roles and structured workflows~\cite{metagpt}. MedAgents studies collaborative medical reasoning~\cite{medagents}, and MDAgents investigates adaptive collaboration for medical decision-making~\cite{mdagents}. These works motivate examining how responsibilities are allocated, but their architectures and evaluations should not be assumed to describe Aegis. The project documentation calls Aegis MetaGPT-based; the available excerpts show a central controller, helper functions, model-assisted validation, and rule-based scoring. A complete executable agent configuration and a controlled comparison with simpler orchestration are absent.

Medical agent benchmarks also distinguish generating a plausible answer from completing an action in a software environment. MedAgentBench evaluates medical LLM agents in a virtual electronic health record environment~\cite{medagentbench}. Aegis has related concerns at the application boundary: a saved answer must belong to the intended session, a report must refer to the correct user, and a message must reach the intended conversation. This connection motivates workflow-level evaluation, but Aegis has not been run on MedAgentBench or any comparable public benchmark. Its contribution remains the documented integration of these functions within one prototype.

\subsection{Retrieval-supported explanation and reporting standards}
Retrieval-augmented generation (RAG) combines retrieved information with language generation~\cite{rag}. Xiong et al. benchmarked RAG for medicine~\cite{medrag}, while Self-RAG investigated retrieval and generation with self-reflection~\cite{selfrag}. Aegis uses a project-maintained lung cancer knowledge resource for explanatory responses. It does not document the experimental configurations or self-reflection mechanisms of these external methods. In particular, the existence of a retrieval step does not establish that the chosen passages are relevant or that the answer accurately represents them.

Evaluation must consequently separate retrieval from generation. RAGAs addresses automated evaluation of retrieved context and generated answers~\cite{ragas}. Liu et al. also show that language-model performance can depend on where relevant information appears in a long context~\cite{lostmiddle}. These findings motivate checking source selection, context assembly, and answer support separately. They do not establish a measured failure rate for Aegis. No retrieval-quality, citation-faithfulness, or answer-safety study is supplied for its current knowledge resource.

Reporting guidance provides a framework for the evidence still needed. TRIPOD+AI addresses clinical prediction models~\cite{tripodai}; TRIPOD-LLM addresses studies using large language models~\cite{tripodllm}. DECIDE-AI concerns early clinical evaluation of AI decision support~\cite{decideai}, and CONSORT-AI concerns clinical trial reports~\cite{consortai}. FUTURE-AI addresses trustworthy development and deployment across the healthcare AI lifecycle~\cite{futureai}. We use these works to frame future evaluation and transparent reporting, not to claim that Aegis has undergone their corresponding study designs or satisfies their checklists.

\section{System Architecture}
\begin{figure}[htbp]
\centering
\includegraphics[width=.87\linewidth]{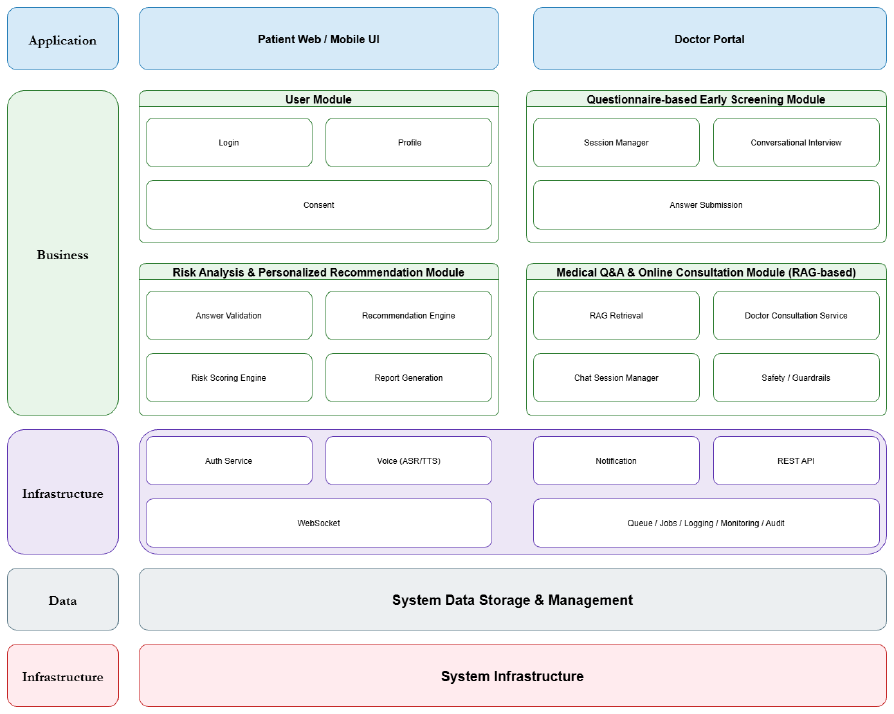}
\caption{Layered design of Aegis, retained from the project report. The diagram summarizes the intended organization of application, business, shared-service, data, and infrastructure components. It is an architecture description, not evidence that every depicted service has been independently verified.}
\label{fig:architecture}
\end{figure}

The prototype comprises four business modules: user management, questionnaire-based assessment, risk analysis and recommendations, and medical question answering with online consultation. Figure~\ref{fig:architecture} preserves the project's layered design. Browser interfaces provide patient-side and doctor-side views; backend services control sessions, verify identity, invoke language-model functions, and persist records. Firebase Authentication, Firestore, and Storage provide identity, structured-record, and file services, respectively.

The main application embeds an independently running questionnaire interface through an iframe and backend proxy. The screening service owns the conversational session and returns the next question or a completed result. The main application then exposes saved reports and follow-up functions. A separate Node.js/Express service handles health-information questions through an \texttt{/api/chat} endpoint. This service boundary allows explanatory dialogue to remain distinct from the assessment state.

The term \emph{multi-agent} denotes the project's functional decomposition. The available implementation excerpts show a central questionnaire manager, session updates, model-assisted validation, and rule-based scoring. They do not establish that each business module is an autonomous language-model agent, or that agents use independent models, negotiation, or consensus. Those stronger interpretations are outside the evidence available for this paper.

\section{Conversational Assessment and Follow-up}
\subsection{Session state and answer handling}
Each questionnaire attempt is organized around a session identifier. The documented session state includes the base question set, current question index, response history, dynamically generated follow-up questions, and confirmed risk factors. The interface shows progress and supports text or voice responses. Although the requirements include pause and resume behavior, test cases TC-16 and TC-17 explicitly state that progress and session recovery after refresh or leaving the page are not supported in the current version. The documented expectation is a clean restart without corrupting stored data.

Responses enter through \texttt{POST /api/intelligent\_questionnaire/reply}. The handler locates the session, checks its validity, and delegates answer processing to the questionnaire manager. The manager updates the answer history and question position and determines whether to ask a further question or complete the session. Responses distinguish a next-question state from a final-result state. The implementation also accommodates question objects and legacy string representations, keeping formatting changes separate from the client-facing interaction.

\begin{figure}[htbp]
\centering
\includegraphics[width=.92\linewidth]{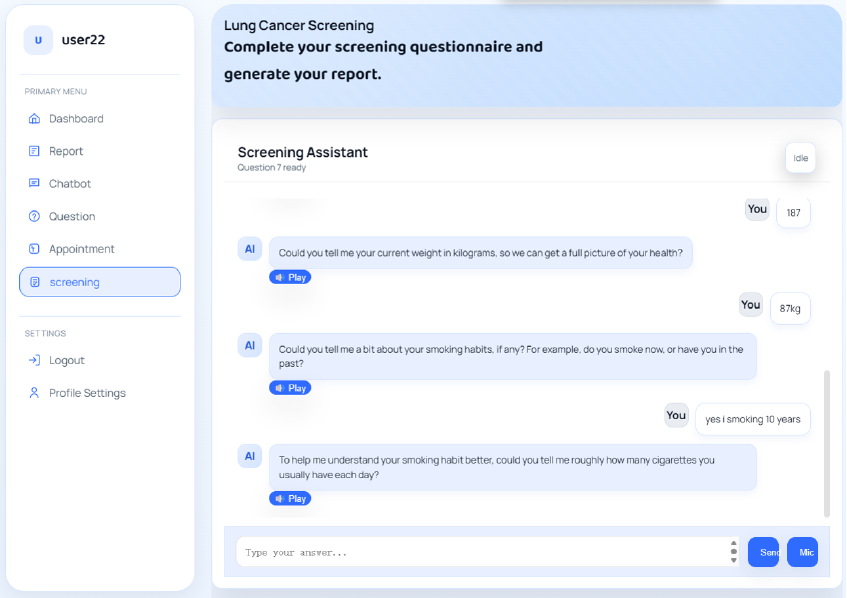}
\caption{Questionnaire interface retained from the project report, showing the conversational collection of self-reported information. The displayed exchange is an interface example, not a clinically validated assessment or evidence of predictive accuracy.}
\label{fig:questionnaire}
\end{figure}

Language-model validation checks whether an answer is relevant, empty, clearly invalid, or a request to repeat earlier information. The prompt uses a permissive policy for ordinary short responses while rejecting clearly unrelated content. Invalid responses lead to a clarification or repeated question; accepted responses update the session context. Dynamic follow-up questions can supplement the base questionnaire when further information is needed. Figure~\ref{fig:questionnaire} shows the conversational interface.

This validation step concerns input handling. It cannot determine whether a user recalls an exposure accurately, understands a symptom correctly, or provides truthful information. Likewise, the screenshots and code excerpts do not quantify whether adaptive questions improve completeness over a fixed questionnaire. Such comparisons require recorded interactions and an independently defined completeness criterion.

\subsection{Risk categorization and report generation}
After collection, the service consolidates answers and risk-related fields into a structured representation. The assessment module checks completeness and applies predefined rules. The report describes factors including age, smoking, passive smoke exposure, occupational exposure, family history, and self-reported symptoms. Its representative code accumulates contributions from factor-specific functions and records explanatory factors alongside the aggregate score.

The system produces a low, medium, or high category, a summary of contributing factors, and suggested follow-up information. Reports are stored as structured records and downloadable PDF files, linked to both the user and the originating session. Historical reports support review and comparison. A change in a displayed category is a change in the prototype's recorded output; without clinical validation it cannot be interpreted as a measured change in disease probability.

The complete scoring functions, category thresholds, and their derivation are not provided in the report. We do not reconstruct them from the examples. This limits reproducibility and prevents a clinical interpretation of category boundaries. In particular, a low category must not be presented as a basis for ruling out disease or withholding clinically indicated assessment.

\subsection{Information support and doctor interaction}
The Q\&A service validates a request, retrieves content from a project-maintained medical knowledge website, includes relevant snippets in the model context, and returns a structured response. The source resource covers lung cancer background, risk factors, screening procedures, and health-management information. System prompts describe an explanatory assistant and discourage diagnostic conclusions. The implementation description also includes external-call timeouts, response-format checks, and error messages.

\begin{figure}[htbp]
\centering
\includegraphics[width=.92\linewidth]{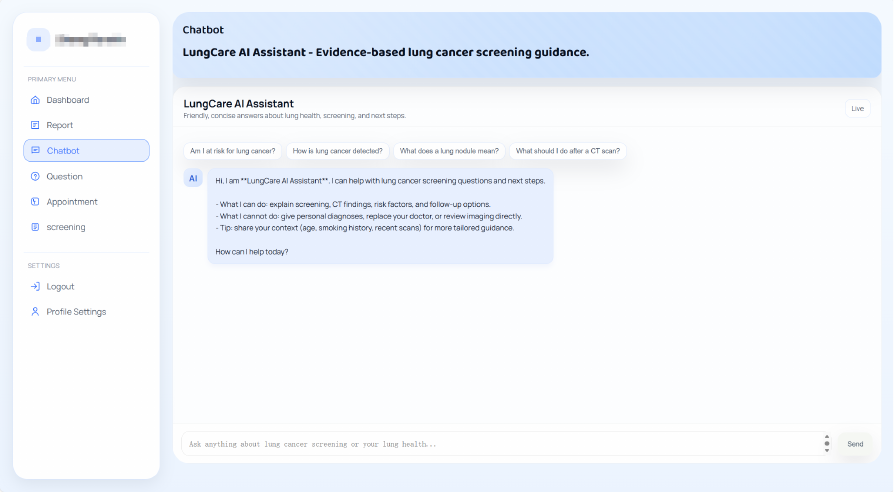}
\caption{Health-information assistant interface retained from the project report. The example shows the explanatory entry point and its non-diagnostic framing. Displayed wording describes the interface's intended role and does not establish the accuracy or safety of generated responses.}
\label{fig:qa}
\end{figure}

Medical disclaimers and prompts are design controls, not guarantees of safe answers. The report does not identify the complete knowledge corpus, its review process, the retrieval configuration, or the model settings. The question-answering module is not used as the documented risk-scoring function. For complex questions or elevated prototype categories, the workflow directs users toward doctor consultation. Availability and timeliness of a clinician response remain external to the software.

Doctor-side views support report inspection, consultation queues, appointments, and risk-status review. Chats and messages retain participant identities, timestamps, and session status. The documented data model also includes CT studies and analysis-result records. However, the report is inconsistent about whether automated imaging inference is implemented or outside scope, and the implementation chapter explicitly states that a full report sign-off workflow is not yet complete. We retain these interfaces as supplementary design material without claiming a validated imaging or sign-off capability.

\section{Implementation and Data Organization}
The browser client uses HTML, CSS, and JavaScript. Python services implement questionnaire control, risk processing, and report generation; Node.js/Express serves the explanatory dialogue endpoint. The report describes cloud deployment with Python and Node.js runtimes but does not supply pinned dependency versions, a reproducible environment, or an evaluation hardware configuration. Its suggested device specifications are design guidance rather than benchmark conditions.

Authentication is based on Firebase ID tokens. Protected backend operations verify identity and make the resulting user identifier available to request handlers. A representative \texttt{/api/whoami} endpoint returns identity context, while \texttt{/api/health/firebase} checks database availability. Shared helpers initialize Firebase services and reuse clients. Role information is retrieved from a user record to distinguish ordinary users from doctors. These mechanisms describe the implementation's access-control structure; no independent security audit or regulatory-compliance assessment is reported.

The data model separates users, screening sessions, answers, risk reports, chats, messages, and appointments. Answer records retain raw input, normalized values, and validation status. Report records reference both user and session identifiers, while binary files are stored separately. Chat metadata supports conversation lists and individual message records preserve the exchange history. This organization makes the intended links between collection, reporting, and consultation explicit. The supplementary schema and interface figures provide further detail.

\section{Functional Verification and Available Evidence}
The project documents unit, integration, and system-level testing. Unit checks concern validation and component behavior; integration checks concern transitions and shared data; system scenarios cover complete patient-side and doctor-side workflows. The test inventory contains 45 cases, summarized in Table~\ref{tab:tests}; the condensed specifications appear in Supplementary Section~S2. Counts in the table are counts of documented specifications, not successful executions.

\begin{table}[htbp]
\centering\small
\caption{Documented functional test inventory. Each original case provides an input, an expected result, and pass/fail criteria. None includes a per-case observed outcome in the supplied report.}
\label{tab:tests}
\begin{tabularx}{\linewidth}{@{}lclY@{}}
\toprule
Area & Cases & Identifiers & Representative coverage\\
\midrule
User management & 11 & TC-01--11 & Registration, consent, authentication, profiles and risk-status display\\
Questionnaire & 8 & TC-12--19 & Session start, voice/text input, clarification, progress, restart and submission\\
User-side reports & 7 & TC-20--26 & History, details, trends, follow-up, download and comparison\\
Doctor-side review & 7 & TC-27--33 & User filtering, records, appointments, imaging linkage and risk adjustment\\
Q\&A and consultation & 12 & TC-34--45 & Disclaimer, information requests, history, messaging and termination\\
\midrule
Total & 45 & TC-01--45 & Functional specifications across five areas\\
\bottomrule
\end{tabularx}
\end{table}

The project's evaluation narrative reports that core flows could be completed, incomplete inputs were intercepted, and information was propagated between questionnaire, report, and doctor-facing views. These are qualitative observations reported by the project team. The supplied material contains no timestamped execution logs, observed-result column, run counts, defect statistics, or independently reproduced tests. Accordingly, no pass rate, reliability estimate, or latency measurement can be derived. The presence of a test specification for an extension also does not establish that the extension was completed.

No clinical reference labels, cohort definition, train/test partition, discrimination metric, calibration analysis, or clinical-outcome evaluation are available. There is also no controlled comparison of adaptive versus fixed questionnaires, agent-based versus simpler orchestration, or retrieved versus ungrounded answers. Requirement-survey screenshots and an interview guide are included in the source material, but participant counts, responses, coding procedures, and quantitative usability results are absent. They are therefore retained as instruments rather than reported as user-study findings.

\section{Evaluation Priorities}
The present evidence supports a description of the prototype and its intended verification scope. A subsequent evaluation should preserve the separation between software correctness, dialogue quality, and clinical interpretation. Table~\ref{tab:evaluation} sets out a proposed evaluation plan. Its entries specify future measurements and do not report completed experiments or estimated performance.

\begin{table}[htbp]
\centering\small
\caption{Proposed evaluation plan. All measurements below remain to be collected.}
\label{tab:evaluation}
\begin{tabularx}{\linewidth}{@{}p{.18\linewidth}YY@{}}
\toprule
Component & Evaluation question & Evidence to collect\\\midrule
Session workflow & Are answers and reports attached to the intended user and attempt? & Per-case observed results, execution traces, authorization checks, restart and duplicate-submission tests\\
Questionnaire dialogue & Does clarification improve information completeness? & Fixed-questionnaire comparison, annotated omissions, correction rates, completion and abandonment\\
Risk rules & What do categories mean for a defined population and outcome? & Published rules and thresholds, reference outcomes, external validation, discrimination and calibration if probabilities are produced\\
Information Q\&A & Are answers supported, relevant, and appropriately bounded? & Corpus snapshot, retrieved passages, clinician-reviewed answers, unsupported claims and unsafe advice\\
Agent organization & Does decomposition improve the same task under a fair comparison? & Matched model and question sets, single-controller baseline, repeated runs, task success, latency and cost\\
User interaction & Can intended users understand and complete the workflow? & Recruitment and consent records, task observations, comprehension checks, accessibility findings\\
\bottomrule
\end{tabularx}
\end{table}

\subsection{Reproducible technical assessment}
The first step is to execute the 45 documented cases against a frozen software version and record actual outcomes separately from expected outcomes. Each record should identify the application version, relevant configuration, input, timestamp, observed response, and any stored-state changes. A clean restart and a recovered session are different behaviors and should remain separate in the results. Tests for report and chat access should use distinct accounts so that cross-user exposure can be assessed directly. These records would allow the qualitative project observations to be examined without interpreting a screenshot as an execution log.

The questionnaire comparison should use the same underlying information requirements for both fixed and adaptive variants. An independent annotation scheme could identify missing fields, contradictions, unnecessary repeated questions, and user corrections. Completion time alone would be insufficient: a shorter session could omit information, while a longer session could collect useful clarification or simply repeat itself. Reporting both completeness and interaction burden would help distinguish these possibilities. Text and voice inputs should be analyzed separately where transcription errors affect the stored answer.

An agent comparison would also require a stable task definition. A central controller and a role-decomposed variant should receive matched cases and comparable model access. Measuring task completion alongside model calls, latency, and cost would reveal whether additional orchestration changes the result or merely the resources used. Repeated runs are needed to characterize output variability. Until such an experiment is performed, the term multi-agent should remain an architectural description rather than a performance claim.

\subsection{Information quality and clinical interpretation}
A Q\&A evaluation should retain the query, retrieved passages, model response, and source version together. This would make it possible to distinguish retrieval failure from unsupported generation. Automated RAG assessment can help organize this analysis~\cite{ragas}, but clinician review is still needed for application-specific judgments about misleading advice and appropriate escalation. Evaluation questions should include ordinary information requests, ambiguous wording, irrelevant inputs, and requests for a diagnosis that exceed the assistant's intended role. These are proposed test categories, not a completed safety benchmark.

Clinical interpretation requires a defined target population and reference outcome before selecting metrics. For the current categorical rules, publishing the exact score construction and thresholds is a prerequisite. If the system is revised to produce a probability, the outcome and time horizon must be specified before calibration has a clear meaning. TRIPOD+AI and TRIPOD-LLM provide relevant reporting guidance for these different components~\cite{tripodai,tripodllm}. Later work involving clinicians in real workflows would require its own protocol and appropriate participant safeguards; DECIDE-AI provides guidance for reporting early clinical evaluation~\cite{decideai}. None of these stages can be inferred from the current functional inventory.

\section{Discussion and Limitations}
The useful engineering property of Aegis is the explicit connection among conversational input, session state, stored reports, and subsequent communication. Separating these responsibilities makes the intended data flow inspectable and creates identifiable points for future measurement. For example, question relevance can be assessed before evaluating the risk categorization, and information quality can be assessed separately from message delivery. These opportunities follow from the design; they are not measured benefits of the current prototype.

\paragraph{Clinical validity.}
The current rules lack reported derivation and outcome-based validation. Self-reported information can be incomplete or incorrect, and rule categories may convey unwarranted confidence if presented without context. Claims of earlier detection, reduced mortality, fewer unnecessary LDCT examinations, cost effectiveness, or reduced clinical workload are not supported by the present evaluation. Any clinical use would require a defined target population, specified prediction task, complete scoring method, and appropriate reference outcomes. Comparison with an established model such as PLCO\textsubscript{M2012} would address a different question from whether the application screens and transitions correctly.

\paragraph{Model behavior and reproducibility.}
The report does not pin the language model, model version, generation settings, MetaGPT configuration, retrieval corpus, or retrieval parameters. It also lacks an ablation isolating the effect of dynamic questions or agent decomposition. The next reproducibility step is to release versioned questionnaire rules, prompts, service configurations, and non-sensitive execution examples. Subsequent experiments should distinguish answer acceptance, information completeness, retrieval support, and clinical relevance. A prompt instructing the model to avoid diagnosis cannot alone demonstrate that it follows this instruction reliably.

\paragraph{Data, accessibility, and deployment.}
The source describes doctor-side content as demonstration material and the imaging examples as obtained from public websites; it does not provide image-level provenance or a license inventory. It also does not provide an ethics-review identifier or consent documentation for the mentioned questionnaires and interviews. No patient-outcome or human-subject findings are asserted here. Those records would be needed for an empirical study involving participants or clinical data. Voice input, responsive interfaces, multilingual support, and role separation remain design features whose accessibility, reliability, and security require direct evaluation. The incomplete sign-off flow and conflicting imaging descriptions further limit claims about deployment readiness.

\section{Conclusion}
Aegis integrates a conversational questionnaire, rule-generated risk information, persistent reports, retrieval-supported explanations, and doctor-facing communication in a web prototype. The available artifacts document its architecture and a 45-case functional test scope. They do not establish predictive accuracy, clinical utility, or an advantage of multi-agent orchestration. A reproducible release and prospectively specified technical and clinical evaluations are the next steps toward assessing whether this workflow is useful beyond the demonstration setting.

\paragraph{Author contributions.}
All four authors contributed equally and share first authorship. Han Bowen developed the questionnaire and screening workflow; Cao Chengwei developed medical Q\&A and online consultation; Zhang Runxin developed risk analysis and recommendations; Jiang Yuyang developed user management. The team contributed jointly to integration, testing, and documentation. Han Bowen is the corresponding author.

\paragraph{Acknowledgements.}
We thank Dr. Justina John for supervision and feedback on the course project from which this manuscript was developed.

\label{main-end}

%% file: supplement.tex
\setcounter{page}{1}
\renewcommand{\thepage}{S\arabic{page}}
\setcounter{section}{0}
\renewcommand{\thesection}{S\arabic{section}}
\renewcommand{\theHsection}{supp.\arabic{section}}
\setcounter{figure}{0}\setcounter{table}{0}
\renewcommand{\thefigure}{S\arabic{figure}}
\renewcommand{\thetable}{S\arabic{table}}
\renewcommand{\theHfigure}{supp.\arabic{figure}}
\renewcommand{\theHtable}{supp.\arabic{table}}

\fancyhead[L]{\small Aegis: supplementary material}
\begin{center}
{\Large\bfseries Supplementary Material}\par\vspace{5pt}
{\large Aegis: A Conversational Prototype for Lung Cancer\\Risk Information and Follow-up}\par\vspace{6pt}
Han Bowen\textsuperscript{\textdagger,*}, Cao Chengwei\textsuperscript{\textdagger}, Zhang Runxin\textsuperscript{\textdagger}, Jiang Yuyang\textsuperscript{\textdagger}\par
{\small School of Computer Sciences, Universiti Sains Malaysia\\
\textsuperscript{\textdagger}Equal contribution and shared first authorship.\quad
\textsuperscript{*}\href{mailto:hanbowen@student.usm.my}{hanbowen@student.usm.my}}
\end{center}
\section{Scope and Interface Contracts}
\small
This supplement preserves technical detail from the course report while separating requirements, implementation descriptions, and observed evidence. All screenshots are historical project artifacts. The report identifies doctor-side records as demonstration content. Interface labels, risk values, aggregate counters, testimonial text, and service claims visible in the images are not study findings, measured system statistics, or endorsements. The figures do not establish completion of every displayed feature.

\begin{center}\small
\centering
\captionof{table}{Condensed module contracts from the system design and implementation chapters.}
\begin{tabularx}{\linewidth}{@{}p{.17\linewidth}YY@{}}\toprule
Module & Inputs & Documented outputs and boundary\\\midrule
User management & Registration, identity tokens, profile edits, consent & Identity context, role-based routing, account and profile records; no clinical inference\\
Questionnaire & User identifier, session identifier, text/voice responses & Next question or completed structured answers; refresh recovery is unsupported in TC-16/17\\
Risk and reports & Structured answers, factor records, report identifiers & Prototype category, factor explanation, stored report and PDF; complete thresholds are unspecified\\
Information Q\&A & User question, retrieved source snippets & Explanatory response or error; no measured medical-safety guarantee\\
Doctor consultation & Authorized records, messages, appointment requests & Conversation and appointment states; full report sign-off is incomplete\\\bottomrule
\end{tabularx}
\end{center}

\paragraph{Data relationships.}
Users link to screening sessions; sessions link to answers and risk reports. Reports retain user and session references and point to stored PDF files. Chats link participants and contain timestamped messages. Appointment records link user, doctor, schedule, and status. CT-study and result entities describe an imaging extension whose implementation scope is inconsistent across the report. These entities should not be interpreted as a validated imaging dataset or a benchmark.

\paragraph{Implementation coverage.}
The report contains representative authentication, Firebase initialization, questionnaire processing, answer-validation, factor-scoring, and language-model invocation excerpts. The excerpts are reproduced as images in Supplementary Section~S5; they are incomplete and do not form an executable release. The source does not provide a complete versioned questionnaire, all scoring thresholds, model settings, or a knowledge-corpus snapshot. No missing implementation is inferred from a screenshot.

\paragraph{Document selection.}
The course cover, declarations and signatures, student identifiers, grading rubrics, administrative schedules, and demonstration credentials are omitted from the public manuscript. The original cross-product comparison chart is also omitted because its qualitative rankings are not supported by a shared evaluation protocol. The remaining 42 technical figures are grouped below; the three main-text figures are not repeated. Figure captions identify their original report numbers and qualify incomplete or purely illustrative content.

\normalsize
\clearpage
\section{Functional Test Specifications}
\label{app:tests}
All 45 source case identifiers are retained. The table condenses inputs and expected behavior; it contains \emph{no observed execution outcomes}. TC-16/17 document unsupported recovery, and TC-24 makes notification delivery conditional on implementation. Passing a workflow case would not establish clinical validity.
\small\setlength{\tabcolsep}{4pt}
\begin{longtable}{@{}>{\raggedright\arraybackslash}p{.09\linewidth}>{\raggedright\arraybackslash}p{.20\linewidth}>{\raggedright\arraybackslash}p{.24\linewidth}>{\raggedright\arraybackslash}p{.41\linewidth}@{}}
\caption{Condensed functional specifications from the original report.}\\
\toprule ID & Function & Action/input & Expected behavior or boundary\\\midrule
\endfirsthead
\multicolumn{4}{l}{\textit{Table S2 continued}}\\\toprule ID & Function & Action/input & Expected behavior or boundary\\\midrule
\endhead
\midrule\multicolumn{4}{r}{\textit{Continued on next page}}\\\endfoot
\bottomrule\endlastfoot
TC-01 & Self-registration & Submit registration fields and privacy agreement. & Create an inactive account and issue verification prompt; reject invalid fields or missing consent.\\[3pt]
TC-02 & Proxy registration & Enter another person's information and relationship. & Store the relationship and create an account awaiting verification; reject incomplete fields.\\[3pt]
TC-03 & Email activation & Open a valid activation link. & Activate the account and permit login with correct credentials.\\[3pt]
TC-04 & Login & Submit correct credentials. & Create a session and route to the correct role interface; reject invalid credentials and cross-role access.\\[3pt]
TC-05 & Logout & Log out from an authenticated session. & End the session and block protected pages until login.\\[3pt]
TC-06 & Password recovery & Request a reset for a registered email. & Provide a reset prompt and deliver a usable link. This condenses the delivery requirement, not the source's account-existence disclosure behavior.\\[3pt]
TC-07 & Password reset & Submit a matching new password through a reset link. & Accept the new password and invalidate the previous password; reject mismatches.\\[3pt]
TC-08 & View profile & Open profile settings. & Show the current account's stored information; prevent unauthenticated or cross-user access.\\[3pt]
TC-09 & Edit profile & Edit fields and save. & Validate input and retain saved updates; cancellation must not save edits.\\[3pt]
TC-10 & Privacy agreement & Accept the agreement before continuing. & Record agreement status/time; block continuation when required agreement is absent.\\[3pt]
TC-11 & Risk status & Open the dashboard or report view. & Display a category consistent with the latest recorded assessment; avoid an unsupported category before assessment.\\[3pt]
TC-12 & Start questionnaire & Start a session after login. & Create the session, show the first question, and initialize progress.\\[3pt]
TC-13 & Voice response & Grant microphone access, speak, and stop. & Show a transcription and allow submission; do not record without permission or accept empty transcription.\\[3pt]
TC-14 & Text response & Submit an answer and continue. & Store the answer and advance correctly; reject empty input.\\[3pt]
TC-15 & Clarification & Provide a vague response and then clarify. & Trigger a follow-up, save clarification, and continue the session.\\[3pt]
TC-16 & Progress display & Inspect progress during a session. & Match progress to answered questions within the session. The source states that progress persistence after refresh is unsupported.\\[3pt]
TC-17 & Leave or refresh & Leave or refresh during the questionnaire and return. & Current limitation: session recovery is unsupported and a restart may occur. Permit a clean restart without crashes or inconsistent duplicate records.\\[3pt]
TC-18 & Submit responses & Complete required questions and confirm submission. & Save responses and open the summary; block incomplete submissions and unintended duplicate processing.\\[3pt]
TC-19 & Immediate summary & Open the post-submission summary. & Show the preliminary category, key factors, and recommendations; do not mislabel it as a final report.\\[3pt]
TC-20 & Report list & Open the report page. & Show the user's time-ordered records or an appropriate empty state; do not expose other users' reports.\\[3pt]
TC-21 & Report details & Select a saved report. & Match displayed content to the selected report identifier.\\[3pt]
TC-22 & Risk trend & Open trends with at least two reports. & Match displayed points and comparisons to the saved records.\\[3pt]
TC-23 & Follow-up progress & Open follow-up status. & Show the recorded stage, period, and next-check information consistently.\\[3pt]
TC-24 & Reminder settings & Set a reminder date/frequency and save. & Persist valid settings. The original case makes notification triggering conditional on implementation.\\[3pt]
TC-25 & Report download & Download a selected report. & Produce an openable file matching its category, factors, and date.\\[3pt]
TC-26 & Compare reports & Select reports from two dates. & Compare the intended records and display relevant category/factor differences.\\[3pt]
TC-27 & Doctor user list & Open the assigned-patient view. & Show relevant medium/high-category users only within authorized scope, with a clear empty state.\\[3pt]
TC-28 & Doctor report access & Select a user and open a report. & Display the matching user's record with appropriate authorization.\\[3pt]
TC-29 & Appointments & Open doctor-side appointment records. & Show consistent dates, statuses, and cancellations.\\[3pt]
TC-30 & Historical review & Inspect a user's history and trends. & Show complete recorded history and corresponding trend information.\\[3pt]
TC-31 & Image upload & Upload a valid imaging file for a selected user. & Link a viewable file to the correct user/date; reject invalid formats and report errors.\\[3pt]
TC-32 & Imaging status & Open imaging results after upload. & Show the corresponding result if available, otherwise a pending state; do not mark missing results complete.\\[3pt]
TC-33 & Risk adjustment & Edit a category and save a reason. & Record the adjustment, retain history, and synchronize patient and doctor views. This is a specification, not proof of completed clinical sign-off.\\[3pt]
TC-34 & Start information Q\&A & Open the chatbot and ask a question. & Provide an interactive session and a response.\\[3pt]
TC-35 & Disclaimer agreement & Confirm the displayed disclaimer. & Record confirmation and gate access according to the designed rule.\\[3pt]
TC-36 & Lung cancer information & Ask about a nodule or screening. & Provide relevant explanatory information with non-diagnostic framing; clinical correctness was not separately measured.\\[3pt]
TC-37 & Q\&A history & Open previous conversations. & Show complete, time-ordered records belonging to the user.\\[3pt]
TC-38 & Manage Q\&A history & Delete a record or clear a session. & Persist the intended change and limit it to the current user's records.\\[3pt]
TC-39 & Consultation request & Select the doctor-chat option. & Create a consultation record and display waiting/connection status.\\[3pt]
TC-40 & Doctor chat queue & Open patient chats. & Show the consultation queue, unread indicators, and corresponding statuses.\\[3pt]
TC-41 & Connect to consultation & Select a user and connect/reply. & Update connection status consistently in both interfaces.\\[3pt]
TC-42 & Message exchange & Exchange multiple text messages. & Display both directions in order with timestamps and without loss.\\[3pt]
TC-43 & Examination questions & Ask about examination preparation or appointments. & Save and retain the doctor's explanation for review.\\[3pt]
TC-44 & Follow-up advice & Provide and acknowledge a follow-up suggestion. & Preserve the advice for subsequent review.\\[3pt]
TC-45 & End consultation & End or exit the conversation. & Mark the session completed, stop further sending, and retain the record.\\[3pt]
\end{longtable}\normalsize\setlength{\tabcolsep}{6pt}
\clearpage
\section{Design Diagrams}
The diagrams document requirements and intended structure. Their presence does not establish implementation or evaluation of every branch.
\par
\noindent
\begin{minipage}[t]{.48\linewidth}
\vspace{0pt}\centering
\includegraphics[width=\linewidth,height=.43\textheight,keepaspectratio]{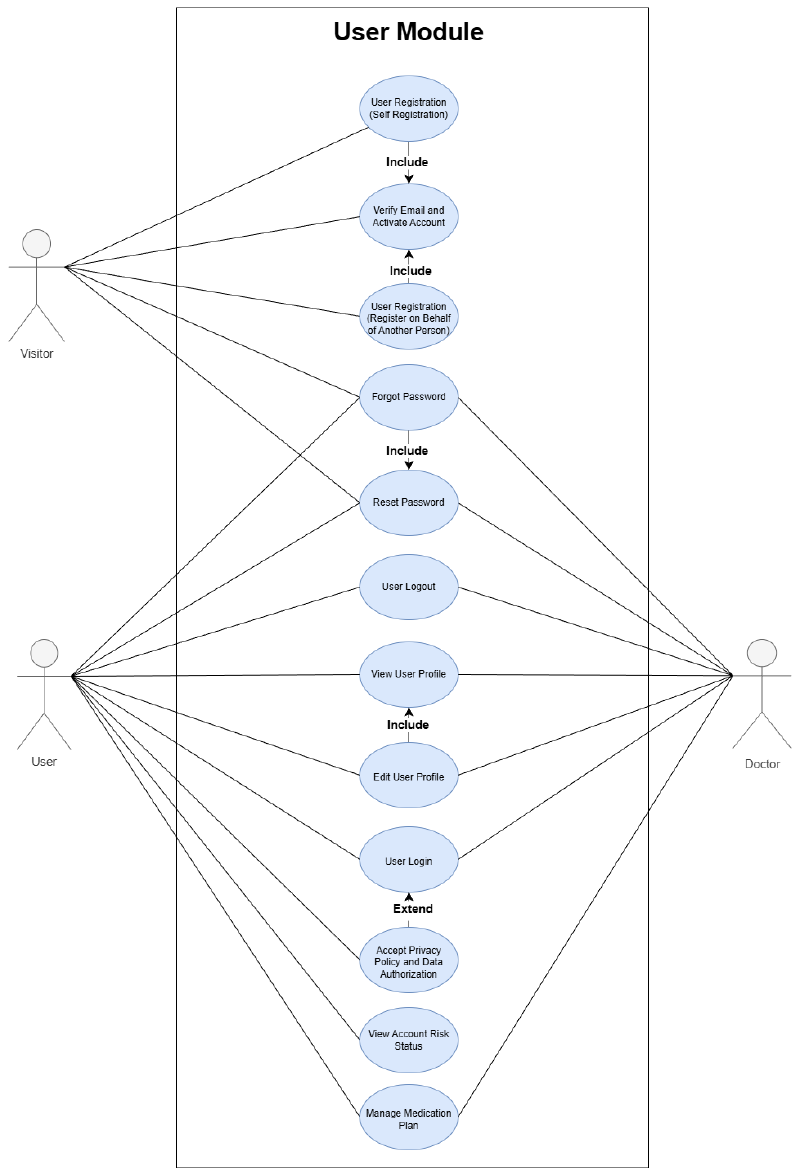}
\captionsetup{font=footnotesize,skip=4pt}
\captionof{figure}{User-management use cases (original Figure 3.1). }
\end{minipage}\hfill
\begin{minipage}[t]{.48\linewidth}
\vspace{0pt}\centering
\includegraphics[width=\linewidth,height=.43\textheight,keepaspectratio]{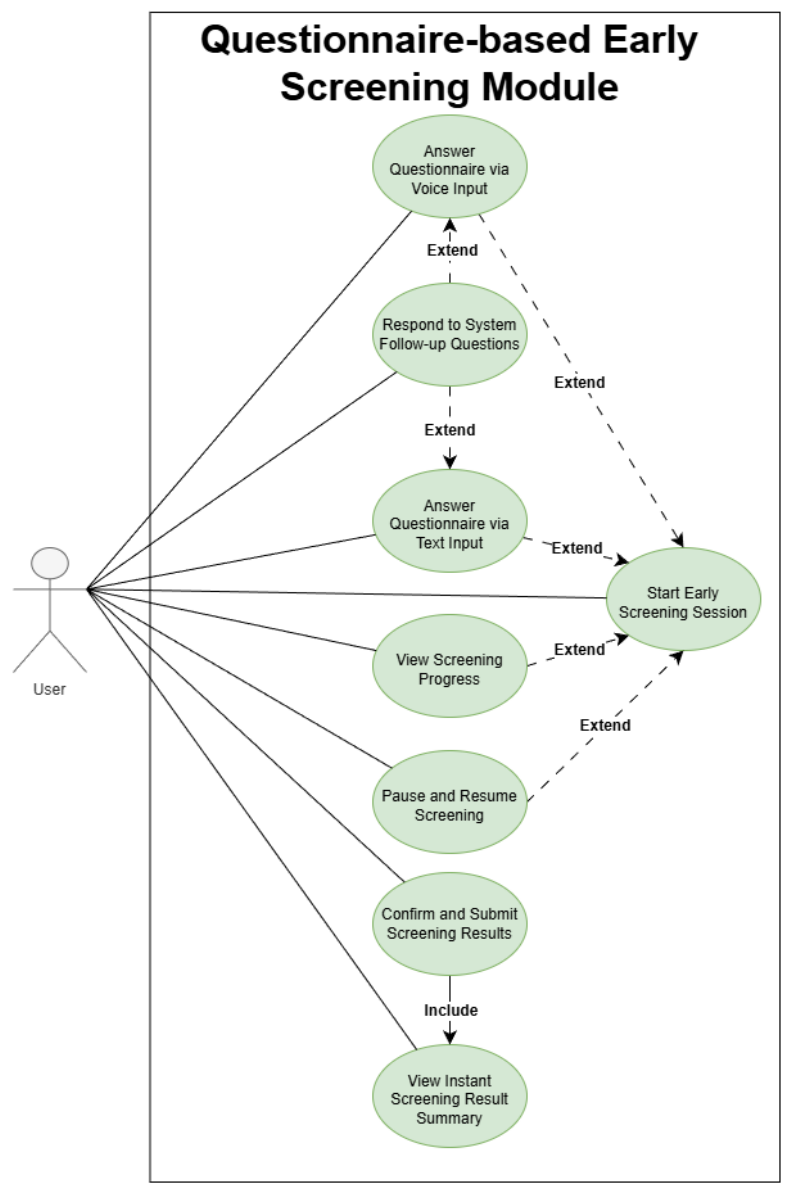}
\captionsetup{font=footnotesize,skip=4pt}
\captionof{figure}{Questionnaire use cases (original Figure 3.2). The diagram includes pause/resume as a requirement; TC-16/17 state that recovery after refresh or leaving the page is unsupported.}
\end{minipage}\par\vspace{9pt}
\par
\noindent
\begin{minipage}[t]{.64\linewidth}
\vspace{0pt}\centering
\includegraphics[width=\linewidth,height=.30\textheight,keepaspectratio]{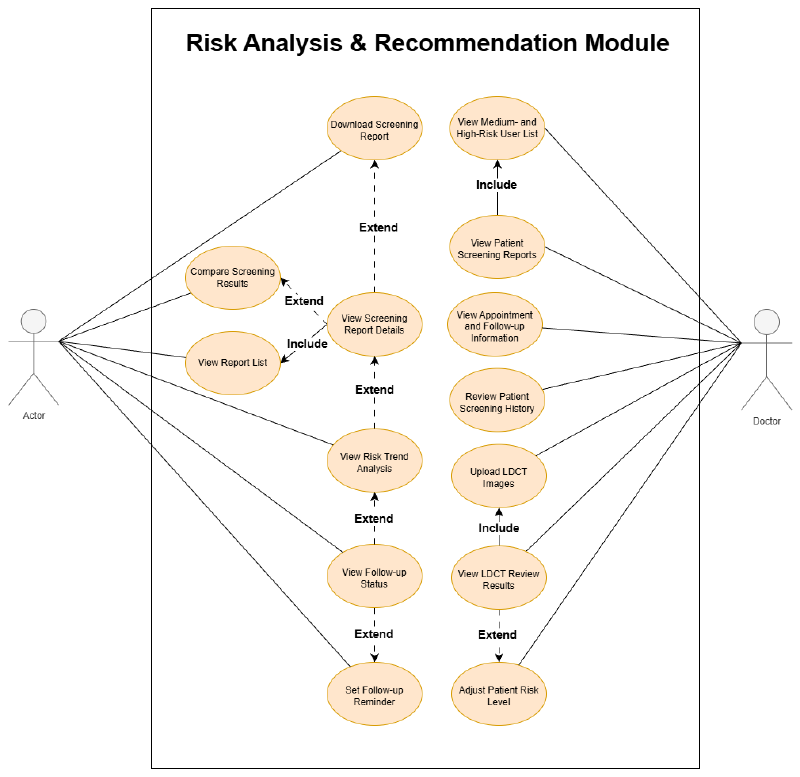}
\captionsetup{font=footnotesize,skip=4pt}
\captionof{figure}{Risk-analysis and recommendation use cases (original Figure 3.3). Risk-analysis requirements, not a validated risk model.}
\end{minipage}\par\vspace{9pt}
\clearpage
\subsection*{Consultation use cases and activity flow}
\par
\noindent
\begin{minipage}[t]{.48\linewidth}
\vspace{0pt}\centering
\includegraphics[width=\linewidth,height=.30\textheight,keepaspectratio]{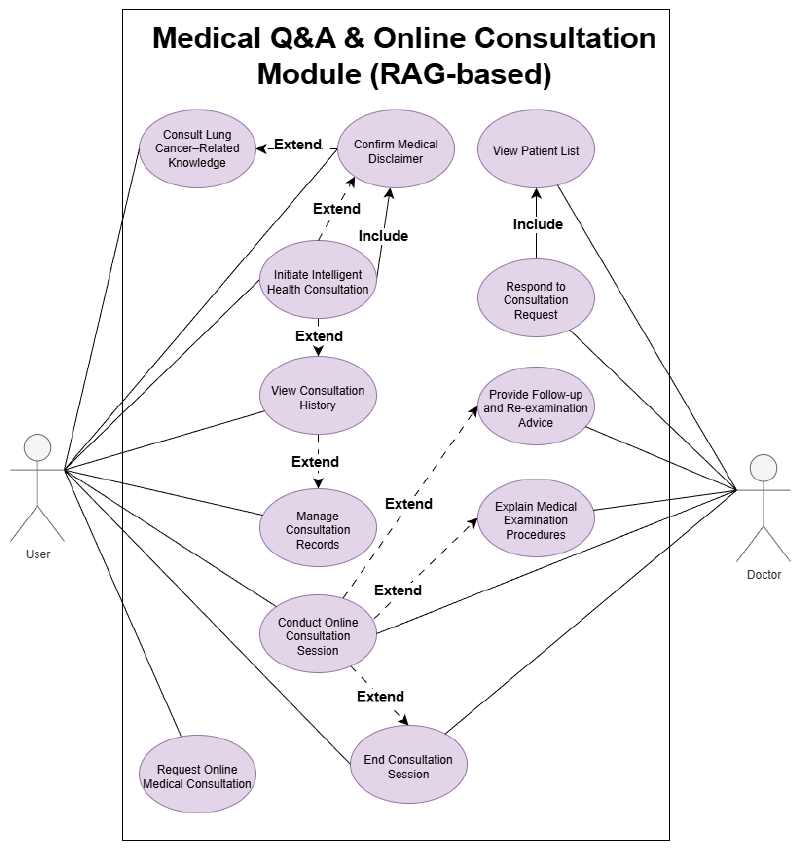}
\captionsetup{font=footnotesize,skip=4pt}
\captionof{figure}{Medical Q\&A and consultation use cases (original Figure 3.4). }
\end{minipage}\hfill
\begin{minipage}[t]{.48\linewidth}
\vspace{0pt}\centering
\includegraphics[width=\linewidth,height=.30\textheight,keepaspectratio]{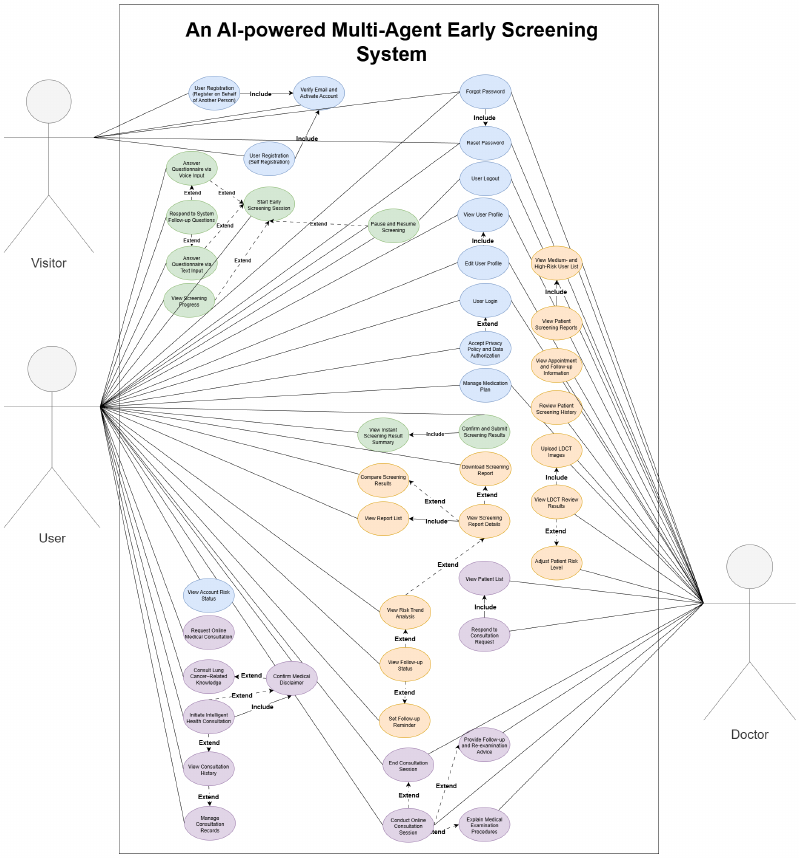}
\captionsetup{font=footnotesize,skip=4pt}
\captionof{figure}{Combined system use cases (original Figure 3.5). }
\end{minipage}\par\vspace{9pt}
\par
\noindent
\begin{minipage}[t]{.70\linewidth}
\vspace{0pt}\centering
\includegraphics[width=\linewidth,height=.43\textheight,keepaspectratio]{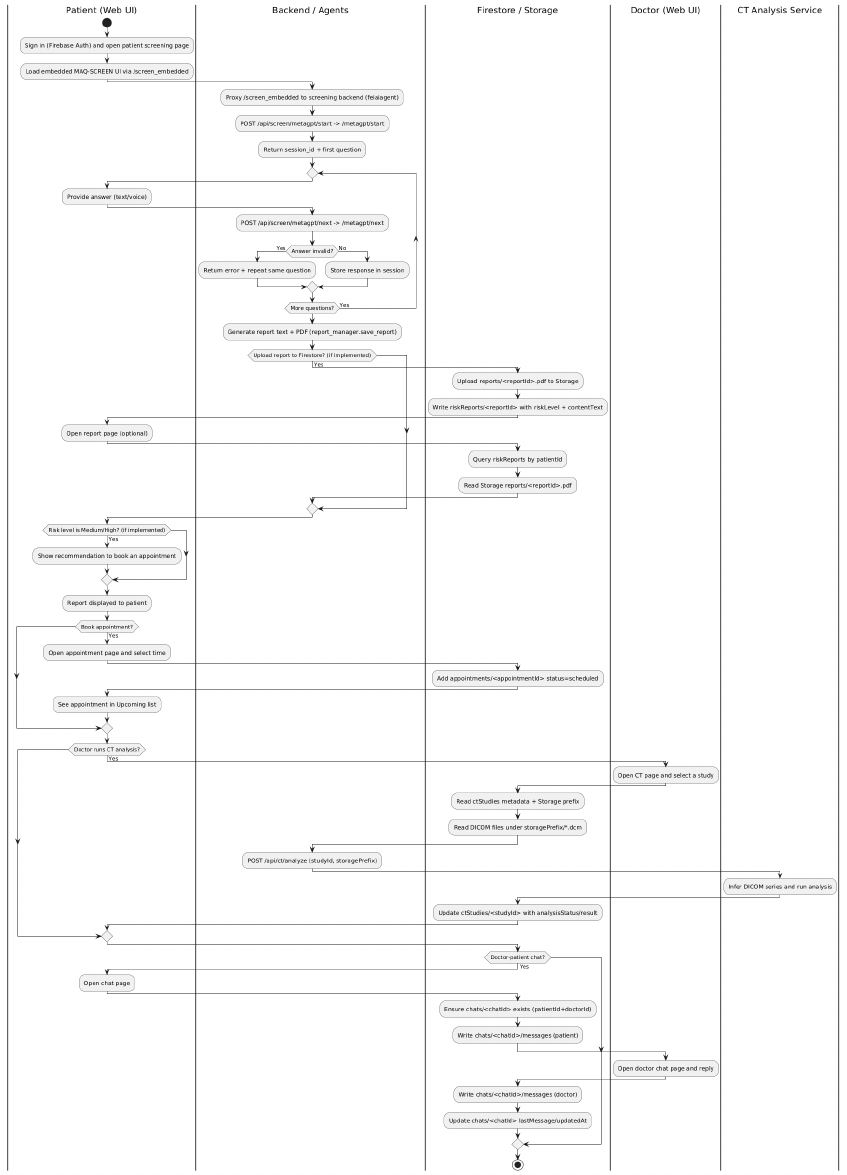}
\captionsetup{font=footnotesize,skip=4pt}
\captionof{figure}{Operational activity diagram (original Figure 3.6). The imaging branch represents the reported design; the report does not consistently establish its implementation status.}
\end{minipage}\par\vspace{9pt}
\clearpage
\subsection*{Screening-to-report sequence}
\par
\noindent
\begin{minipage}[t]{.99\linewidth}
\vspace{0pt}\centering
\includegraphics[width=\linewidth,height=.80\textheight,keepaspectratio]{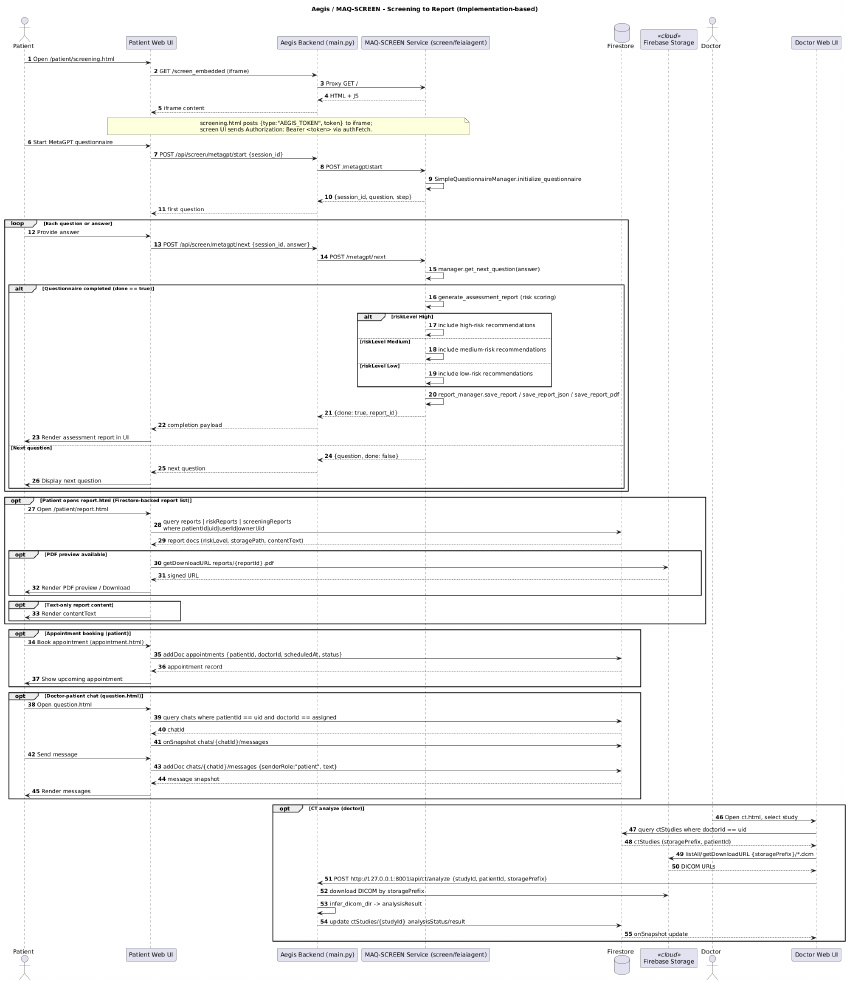}
\captionsetup{font=footnotesize,skip=4pt}
\captionof{figure}{Screening-to-report sequence diagram (original Figure 3.7). The sequence is an implementation description, not an execution trace or measured latency result.}
\end{minipage}\par\vspace{9pt}
\clearpage
\begin{landscape}
\subsection*{Patient-side state description}
\begin{center}
\includegraphics[angle=-90,origin=c,width=.99\linewidth,totalheight=.72\textheight,keepaspectratio]{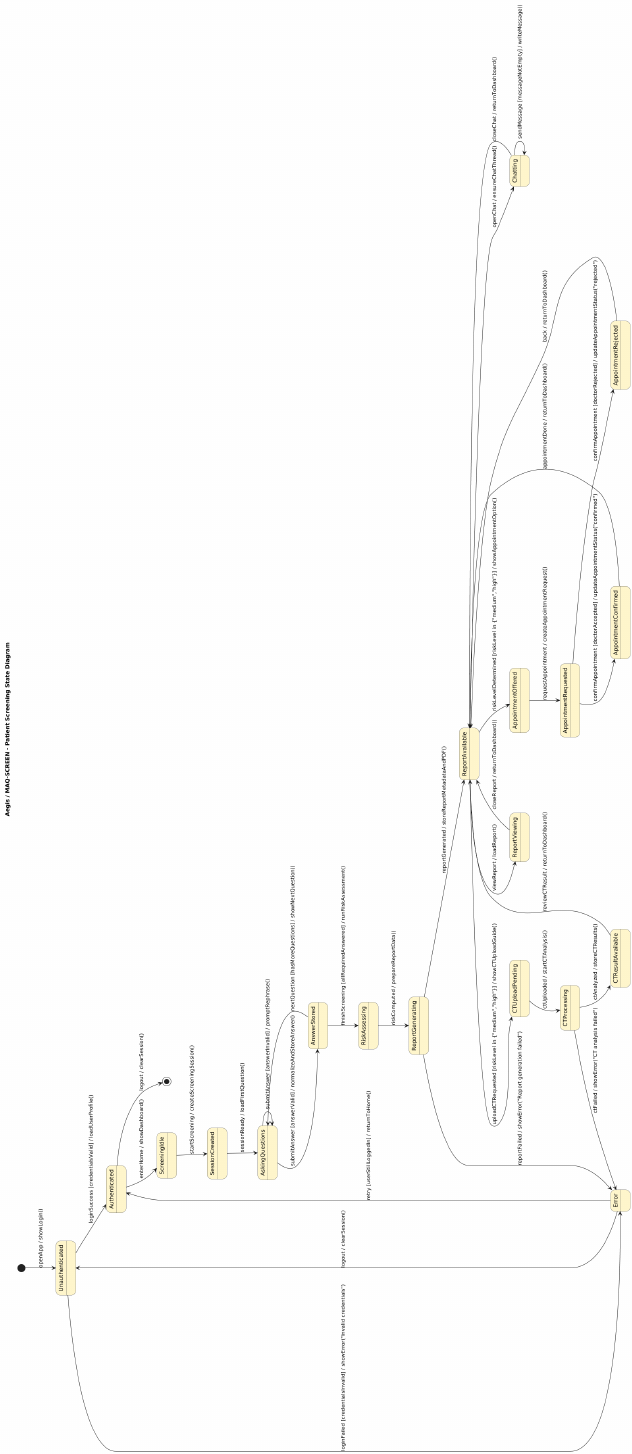}
\captionsetup{font=footnotesize,skip=6pt}
\captionof{figure}{Patient-side state diagram (original Figure 3.8). The original diagram is rotated for readability; its labels and structure are preserved. This design view is not a recorded execution trace.}
\end{center}
\end{landscape}
\clearpage
\subsection*{Architecture and data structure}
\par
\noindent
\begin{minipage}[t]{.52\linewidth}
\vspace{0pt}\centering
\includegraphics[width=\linewidth,height=.22\textheight,keepaspectratio]{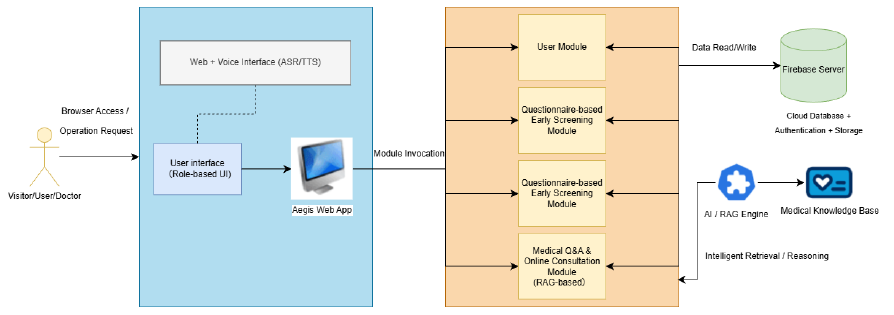}
\captionsetup{font=footnotesize,skip=4pt}
\captionof{figure}{Original high-level architecture (original Figure 4.1). This original diagram repeats the questionnaire label in two module boxes; the corrected four-module description is given in the main text.}
\end{minipage}\hfill
\begin{minipage}[t]{.44\linewidth}
\vspace{0pt}\centering
\includegraphics[width=\linewidth,height=.25\textheight,keepaspectratio]{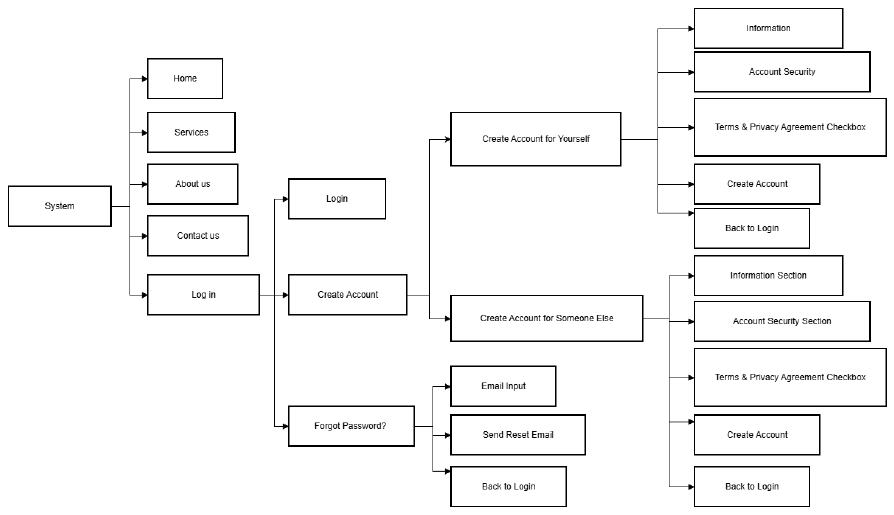}
\captionsetup{font=footnotesize,skip=4pt}
\captionof{figure}{Account-access interface hierarchy (original Figure 4.4). }
\end{minipage}\par\vspace{9pt}
\par
\noindent
\begin{minipage}[t]{.99\linewidth}
\vspace{0pt}\centering
\includegraphics[width=\linewidth,height=.46\textheight,keepaspectratio]{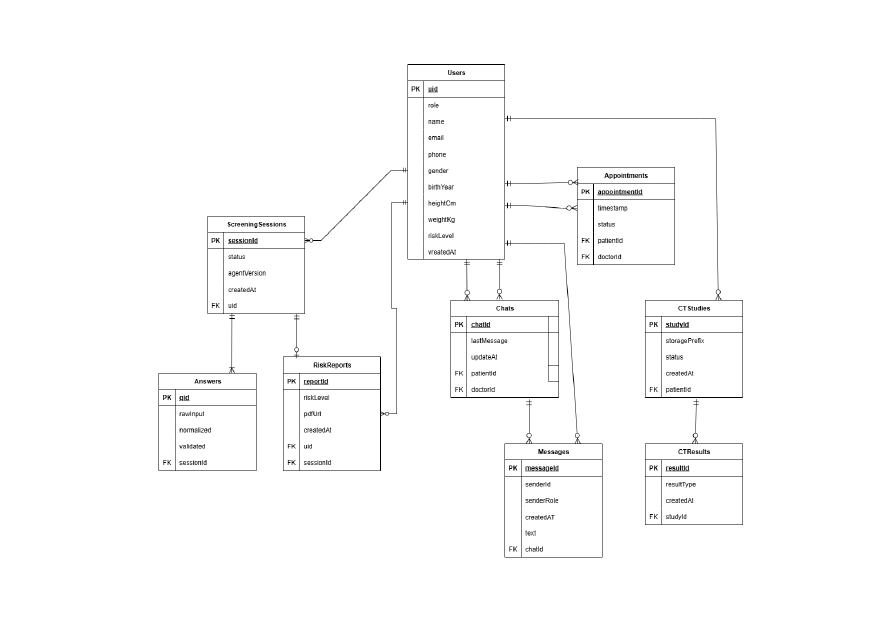}
\captionsetup{font=footnotesize,skip=4pt}
\captionof{figure}{Database and class diagram (original Figure 4.3). Entity and relationship design, including an imaging extension whose completion is not established.}
\end{minipage}\par\vspace{9pt}
\clearpage
\section{Additional Interface Examples}
Screenshots preserve the historical prototype. Service claims, testimonials, counters, scores, and clinical-looking records are illustrative content, not study findings.
\par
\noindent
\begin{minipage}[t]{.48\linewidth}
\vspace{0pt}\centering
\includegraphics[width=\linewidth,height=.20\textheight,keepaspectratio]{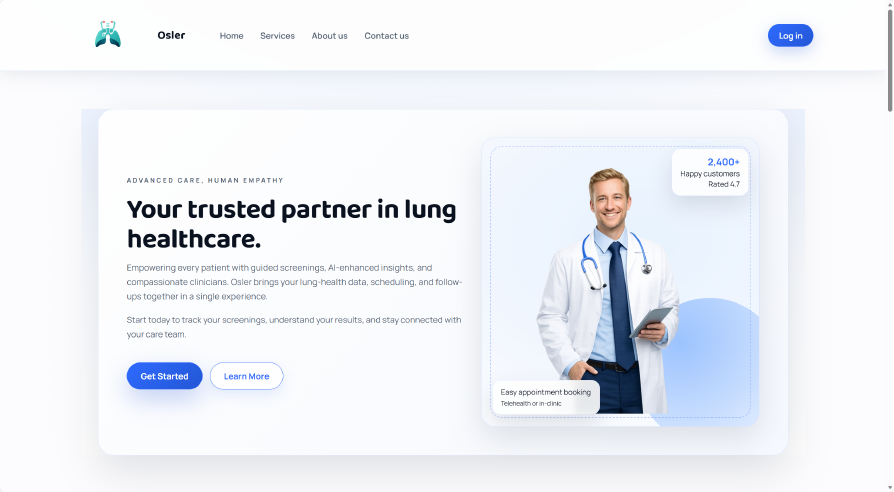}
\captionsetup{font=footnotesize,skip=4pt}
\captionof{figure}{Public home-page interface (original Figure 4.5). Illustrative public-page content.}
\end{minipage}\hfill
\begin{minipage}[t]{.48\linewidth}
\vspace{0pt}\centering
\includegraphics[width=\linewidth,height=.20\textheight,keepaspectratio]{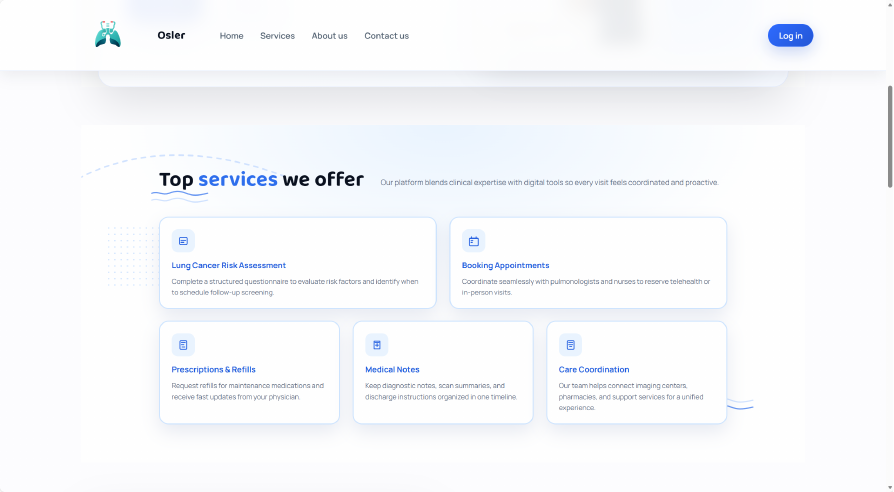}
\captionsetup{font=footnotesize,skip=4pt}
\captionof{figure}{Service overview interface (original Figure 4.6). Illustrative service descriptions.}
\end{minipage}\par\vspace{9pt}
\par
\noindent
\begin{minipage}[t]{.48\linewidth}
\vspace{0pt}\centering
\includegraphics[width=\linewidth,height=.20\textheight,keepaspectratio]{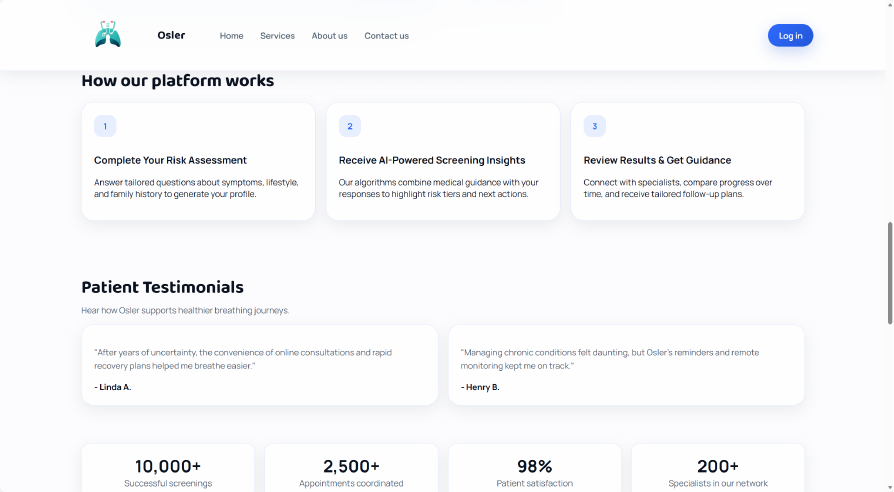}
\captionsetup{font=footnotesize,skip=4pt}
\captionof{figure}{Public workflow and testimonial layout (original Figure 4.7). Testimonials and counters are not verified outcomes.}
\end{minipage}\hfill
\begin{minipage}[t]{.48\linewidth}
\vspace{0pt}\centering
\includegraphics[width=\linewidth,height=.20\textheight,keepaspectratio]{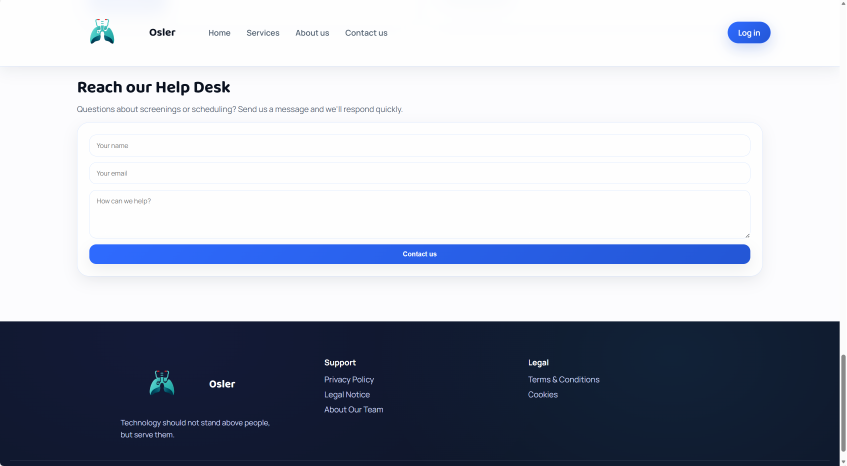}
\captionsetup{font=footnotesize,skip=4pt}
\captionof{figure}{Contact interface (original Figure 4.8). Contact-page design.}
\end{minipage}\par\vspace{9pt}
\par
\noindent
\begin{minipage}[t]{.48\linewidth}
\vspace{0pt}\centering
\includegraphics[width=\linewidth,height=.20\textheight,keepaspectratio]{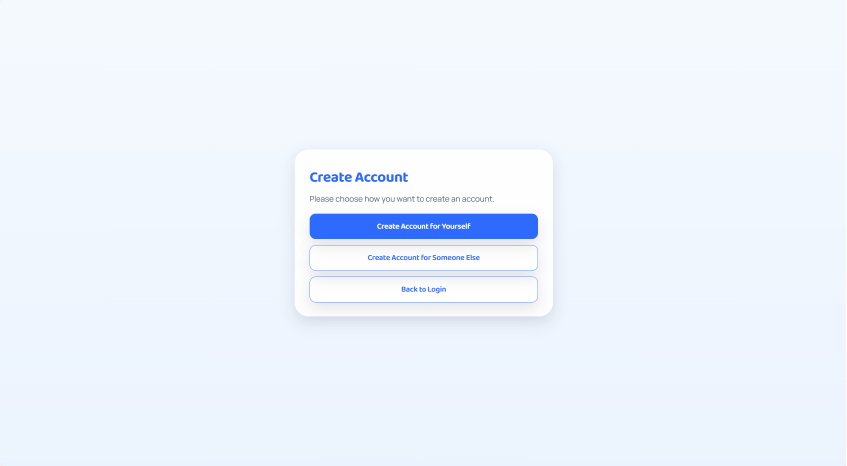}
\captionsetup{font=footnotesize,skip=4pt}
\captionof{figure}{Account-type selection (original Figure 4.9). }
\end{minipage}\hfill
\begin{minipage}[t]{.48\linewidth}
\vspace{0pt}\centering
\includegraphics[width=\linewidth,height=.20\textheight,keepaspectratio]{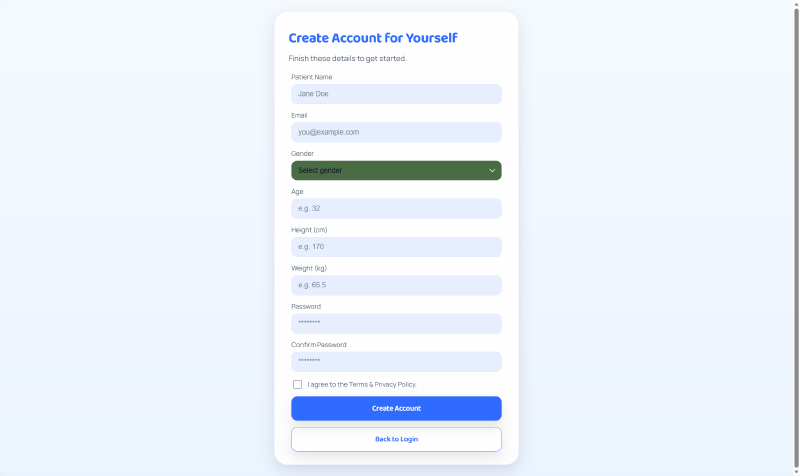}
\captionsetup{font=footnotesize,skip=4pt}
\captionof{figure}{Self-registration interface (original Figure 4.10). }
\end{minipage}\par\vspace{9pt}
\clearpage
\subsection*{Account and patient-side interfaces}
\par
\noindent
\begin{minipage}[t]{.48\linewidth}
\vspace{0pt}\centering
\includegraphics[width=\linewidth,height=.20\textheight,keepaspectratio]{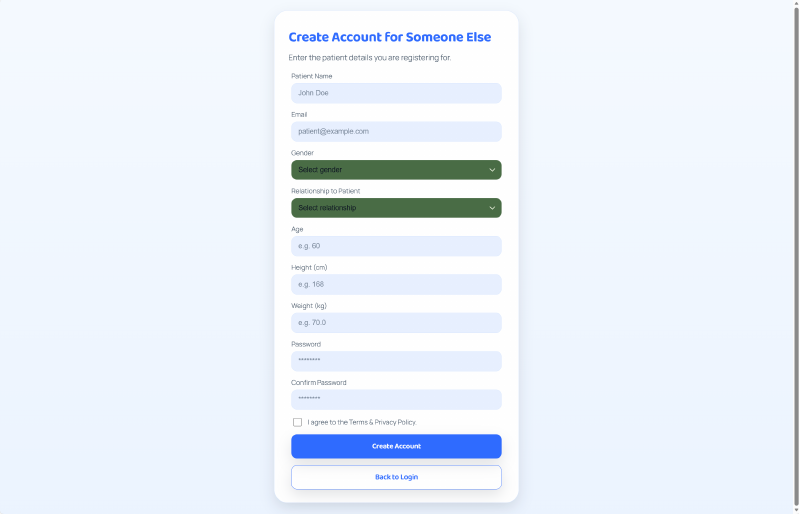}
\captionsetup{font=footnotesize,skip=4pt}
\captionof{figure}{Proxy-registration interface (original Figure 4.11). }
\end{minipage}\hfill
\begin{minipage}[t]{.48\linewidth}
\vspace{0pt}\centering
\includegraphics[width=\linewidth,height=.20\textheight,keepaspectratio]{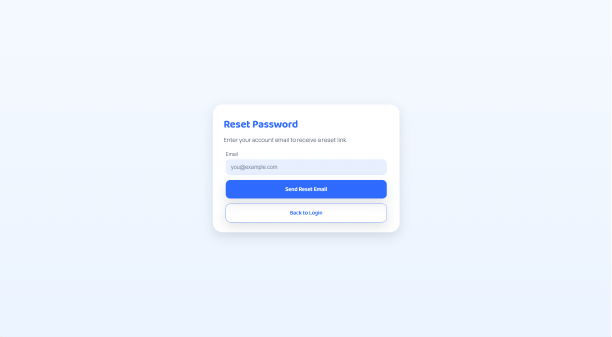}
\captionsetup{font=footnotesize,skip=4pt}
\captionof{figure}{Password-recovery interface (original Figure 4.12). }
\end{minipage}\par\vspace{9pt}
\par
\noindent
\begin{minipage}[t]{.48\linewidth}
\vspace{0pt}\centering
\includegraphics[width=\linewidth,height=.41\textheight,keepaspectratio]{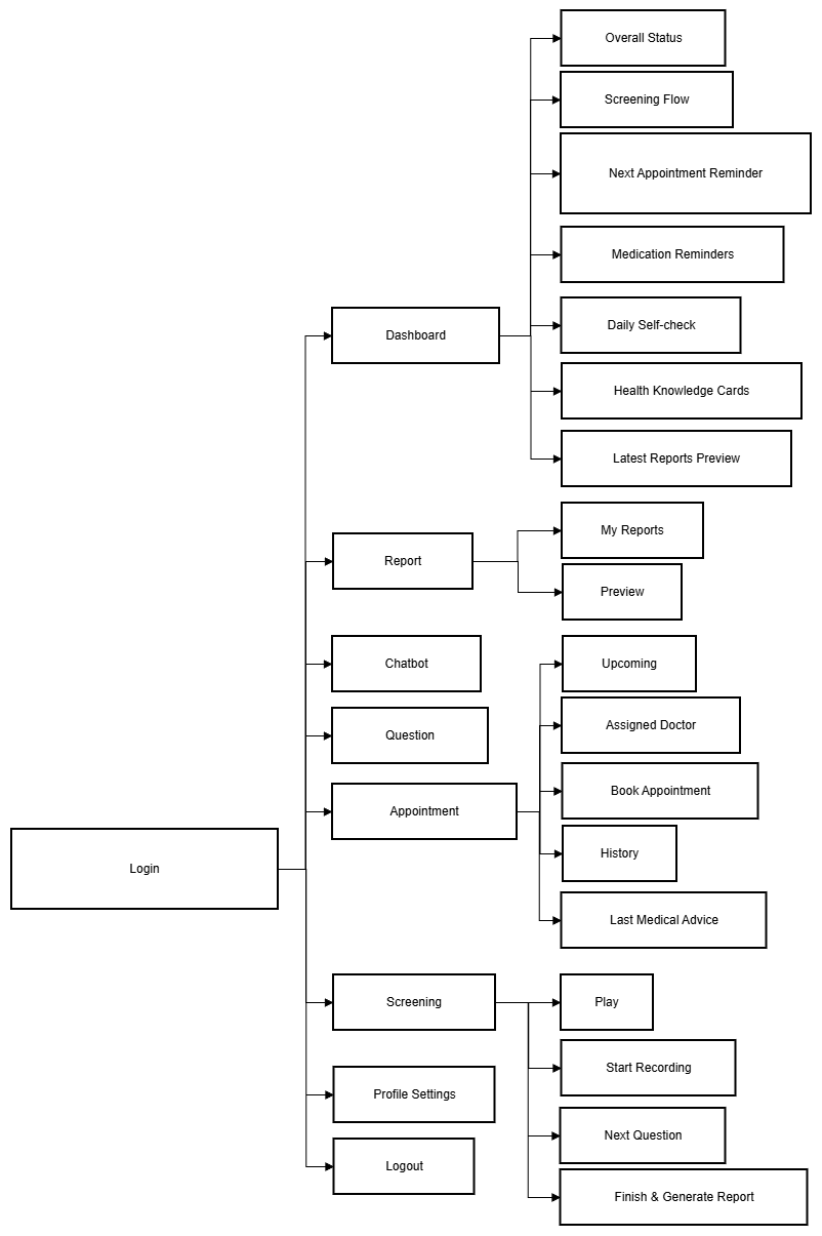}
\captionsetup{font=footnotesize,skip=4pt}
\captionof{figure}{Patient interface hierarchy (original Figure 4.13). }
\end{minipage}\hfill
\begin{minipage}[t]{.48\linewidth}
\vspace{0pt}\centering
\includegraphics[width=\linewidth,height=.41\textheight,keepaspectratio]{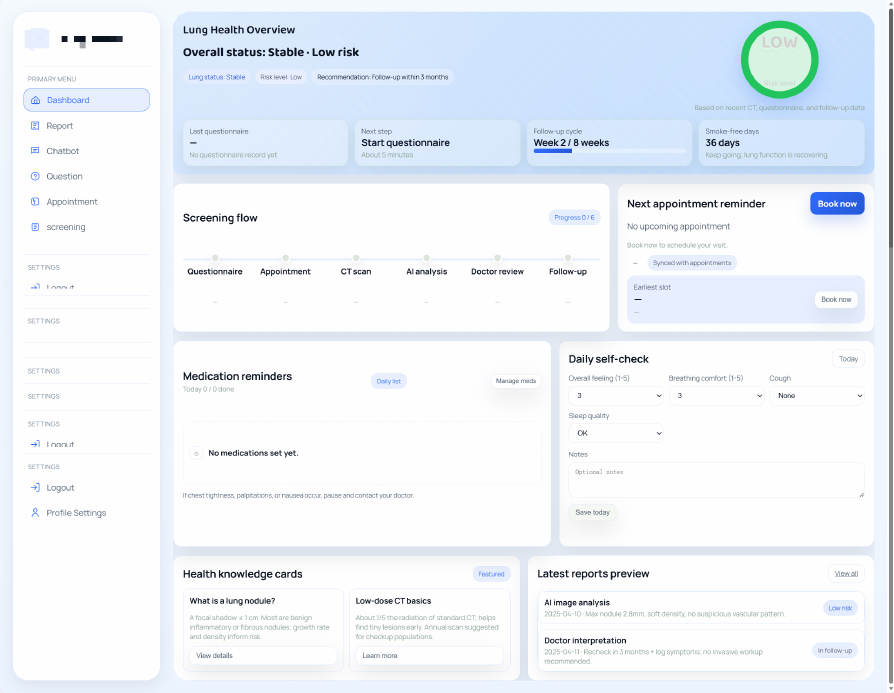}
\captionsetup{font=footnotesize,skip=4pt}
\captionof{figure}{Patient dashboard (original Figure 4.14). Displayed categories, scores, and activity indicators are illustrative; no corresponding clinical outcome data are supplied.}
\end{minipage}\par\vspace{9pt}
\par
\noindent
\begin{minipage}[t]{.61\linewidth}
\vspace{0pt}\centering
\includegraphics[width=\linewidth,height=.20\textheight,keepaspectratio]{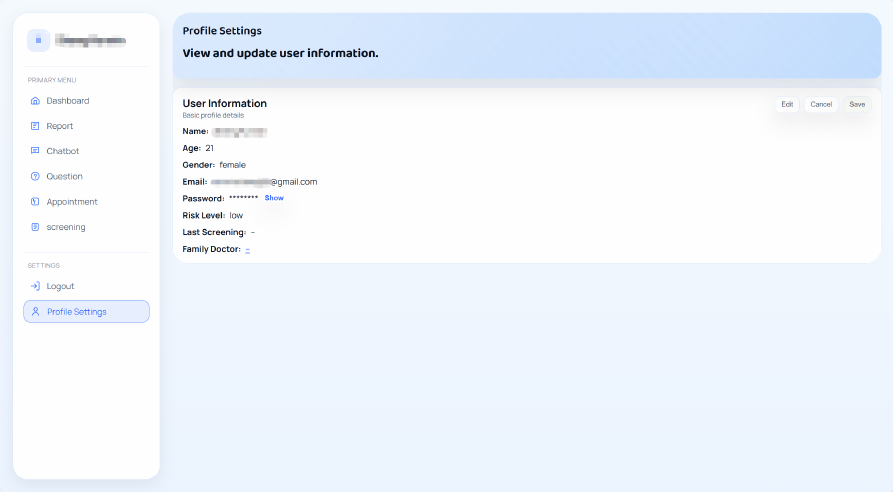}
\captionsetup{font=footnotesize,skip=4pt}
\captionof{figure}{Patient profile interface (original Figure 4.19). }
\end{minipage}\par\vspace{9pt}
\clearpage
\subsection*{Reports, consultation, and doctor overview}
\par
\noindent
\begin{minipage}[t]{.49\linewidth}
\vspace{0pt}\centering
\includegraphics[width=\linewidth,height=.44\textheight,keepaspectratio]{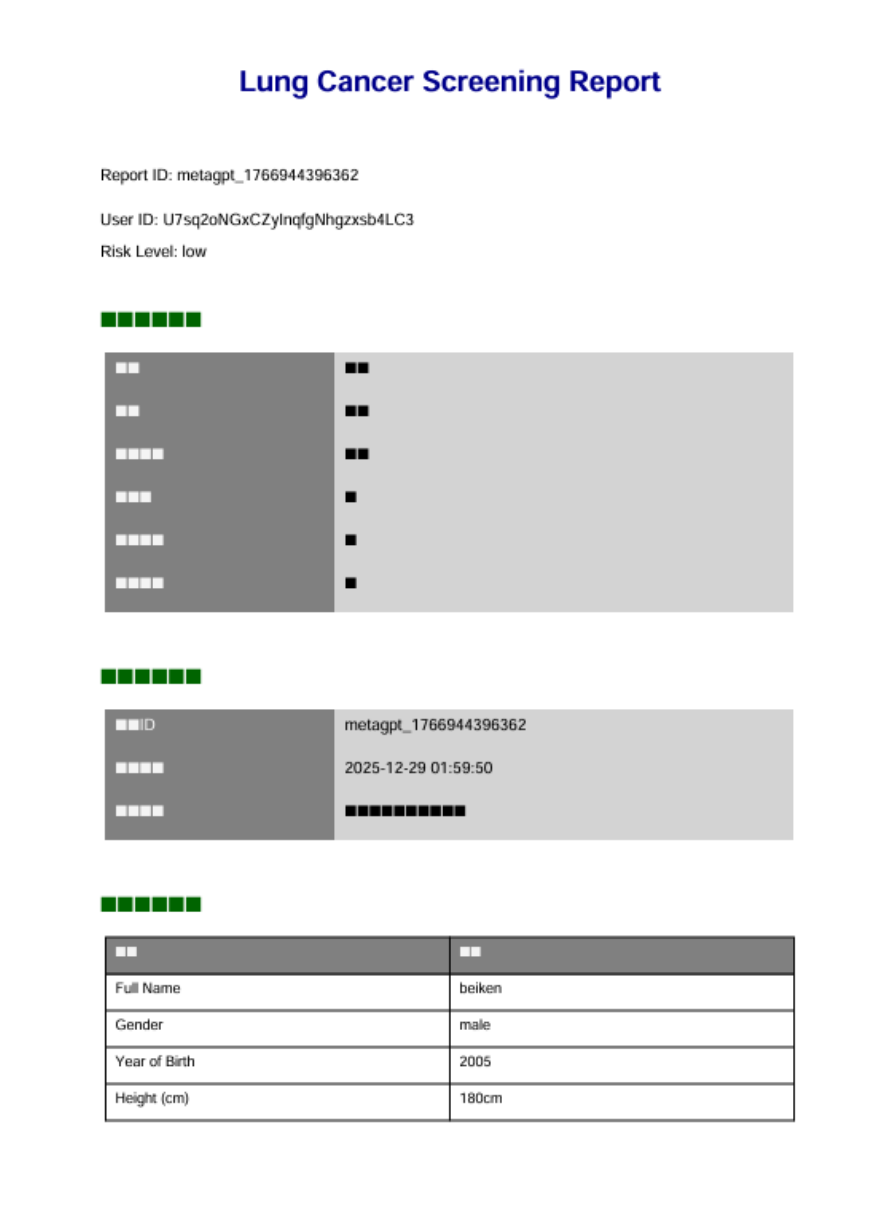}
\captionsetup{font=footnotesize,skip=4pt}
\captionof{figure}{Example generated report (original Figure 4.15). The source image contains missing-glyph boxes in some non-English fields. These source defects are preserved, not reconstructed. This example is not evidence of a validated assessment.}
\end{minipage}\hfill
\begin{minipage}[t]{.47\linewidth}
\vspace{0pt}\centering
\includegraphics[width=\linewidth,height=.44\textheight,keepaspectratio]{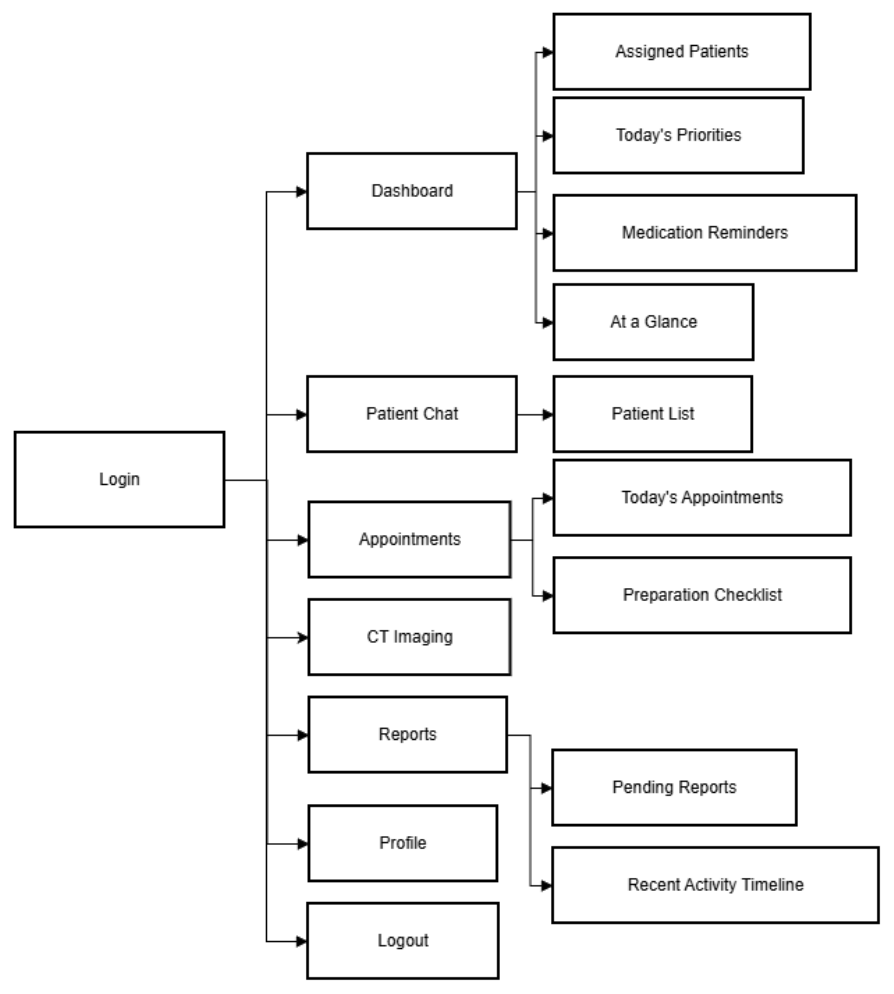}
\captionsetup{font=footnotesize,skip=4pt}
\captionof{figure}{Doctor interface hierarchy (original Figure 4.20). }
\end{minipage}\par\vspace{9pt}
\par
\noindent
\begin{minipage}[t]{.48\linewidth}
\vspace{0pt}\centering
\includegraphics[width=\linewidth,height=.30\textheight,keepaspectratio]{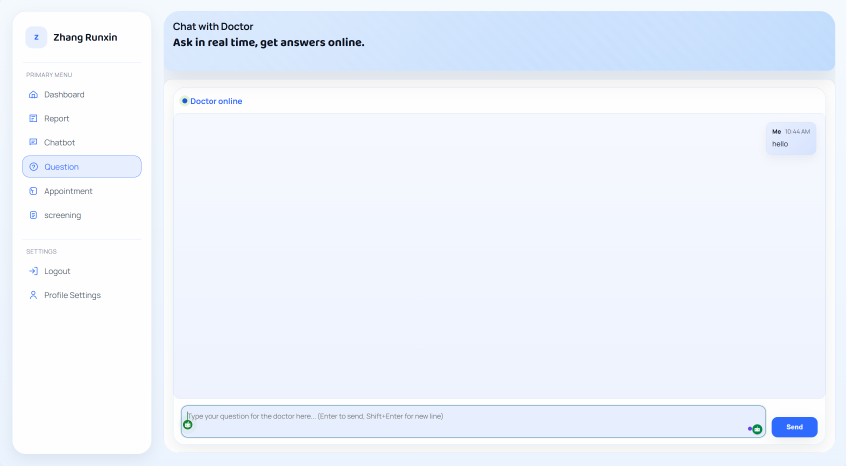}
\captionsetup{font=footnotesize,skip=4pt}
\captionof{figure}{Doctor-consultation interface (original Figure 4.17). }
\end{minipage}\hfill
\begin{minipage}[t]{.48\linewidth}
\vspace{0pt}\centering
\includegraphics[width=\linewidth,height=.30\textheight,keepaspectratio]{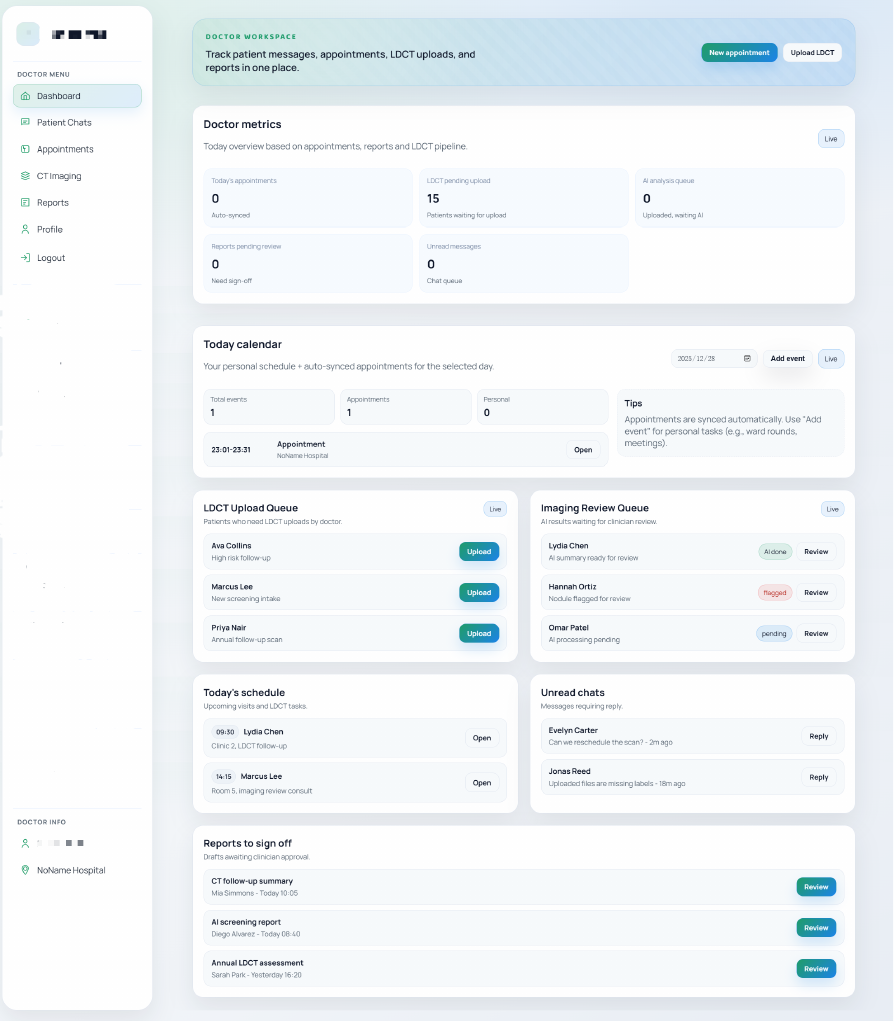}
\captionsetup{font=footnotesize,skip=4pt}
\captionof{figure}{Doctor dashboard (original Figure 4.21). Dashboard counters and clinical-looking records are demonstration content, not measured study results.}
\end{minipage}\par\vspace{9pt}
\clearpage
\subsection*{Doctor-side communication and review}
\par
\noindent
\begin{minipage}[t]{.48\linewidth}
\vspace{0pt}\centering
\includegraphics[width=\linewidth,height=.21\textheight,keepaspectratio]{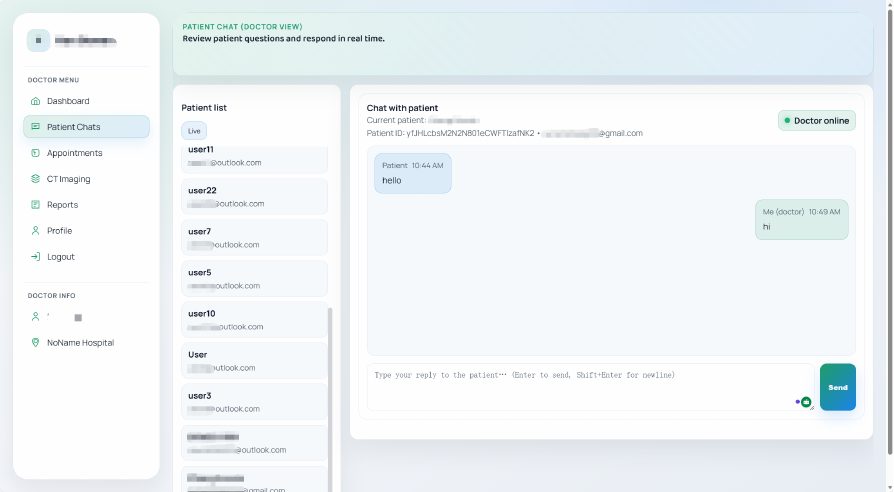}
\captionsetup{font=footnotesize,skip=4pt}
\captionof{figure}{Doctor chat interface (original Figure 4.22). }
\end{minipage}\hfill
\begin{minipage}[t]{.48\linewidth}
\vspace{0pt}\centering
\includegraphics[width=\linewidth,height=.21\textheight,keepaspectratio]{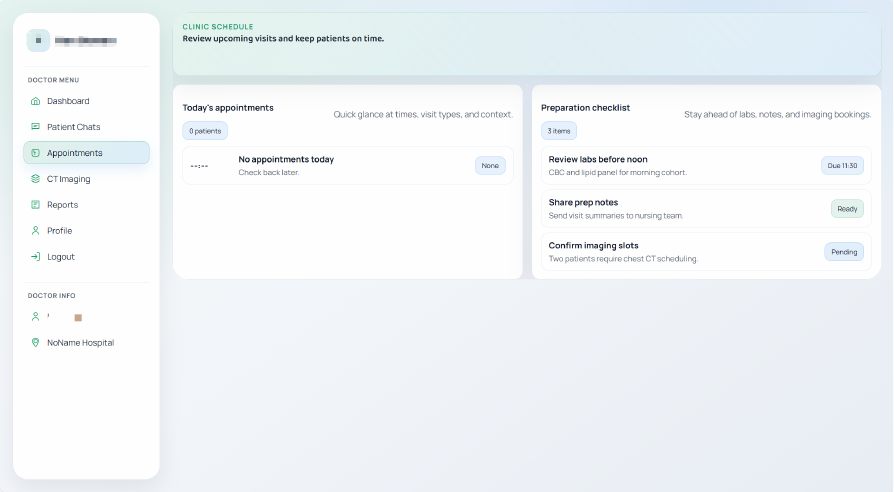}
\captionsetup{font=footnotesize,skip=4pt}
\captionof{figure}{Appointment interface (original Figure 4.23). }
\end{minipage}\par\vspace{9pt}
\par
\noindent
\begin{minipage}[t]{.48\linewidth}
\vspace{0pt}\centering
\includegraphics[width=\linewidth,height=.21\textheight,keepaspectratio]{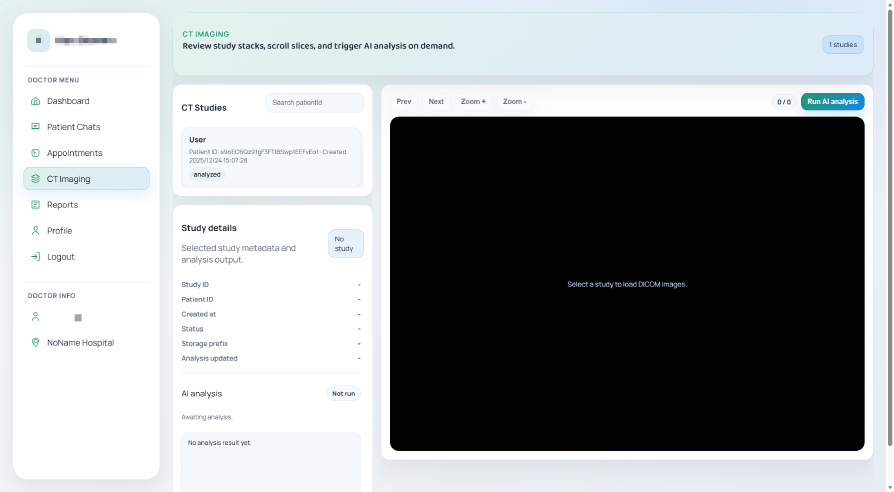}
\captionsetup{font=footnotesize,skip=4pt}
\captionof{figure}{CT-study interface (original Figure 4.24). This interface does not establish that automated CT inference is complete or clinically validated; source descriptions conflict.}
\end{minipage}\hfill
\begin{minipage}[t]{.48\linewidth}
\vspace{0pt}\centering
\includegraphics[width=\linewidth,height=.21\textheight,keepaspectratio]{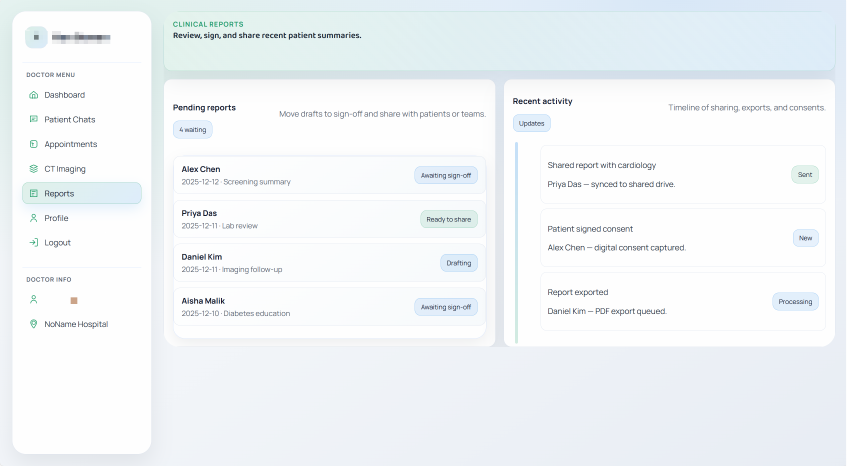}
\captionsetup{font=footnotesize,skip=4pt}
\captionof{figure}{Report-review and sign-off interface (original Figure 4.25). The implementation chapter states that full report sign-off is incomplete despite the displayed review controls.}
\end{minipage}\par\vspace{9pt}
\par
\noindent
\begin{minipage}[t]{.60\linewidth}
\vspace{0pt}\centering
\includegraphics[width=\linewidth,height=.20\textheight,keepaspectratio]{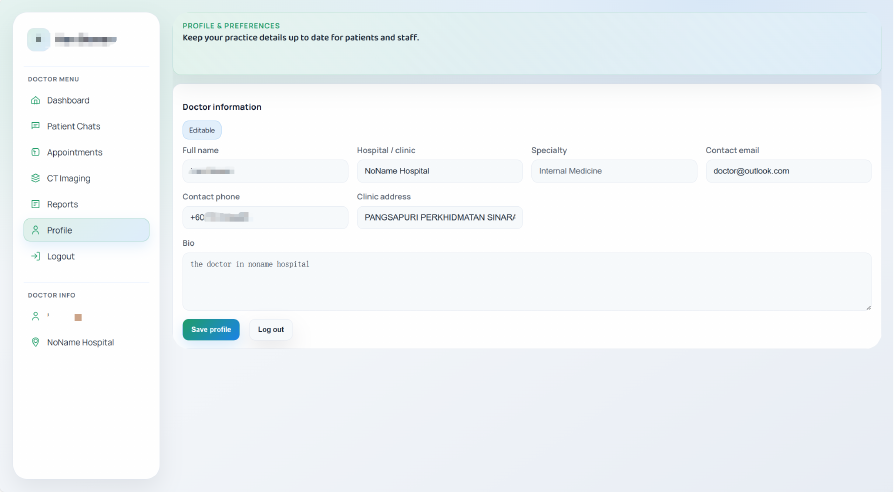}
\captionsetup{font=footnotesize,skip=4pt}
\captionof{figure}{Doctor profile interface (original Figure 4.26). Demonstration profile values.}
\end{minipage}\par\vspace{9pt}
\clearpage
\section{Implementation Excerpts}
These are representative excerpts from the source report. They do not constitute an executable release or supply all dependencies, prompts, scoring functions, and configuration values.
\par
\noindent
\begin{minipage}[t]{.48\linewidth}
\vspace{0pt}\centering
\includegraphics[width=\linewidth,height=.24\textheight,keepaspectratio]{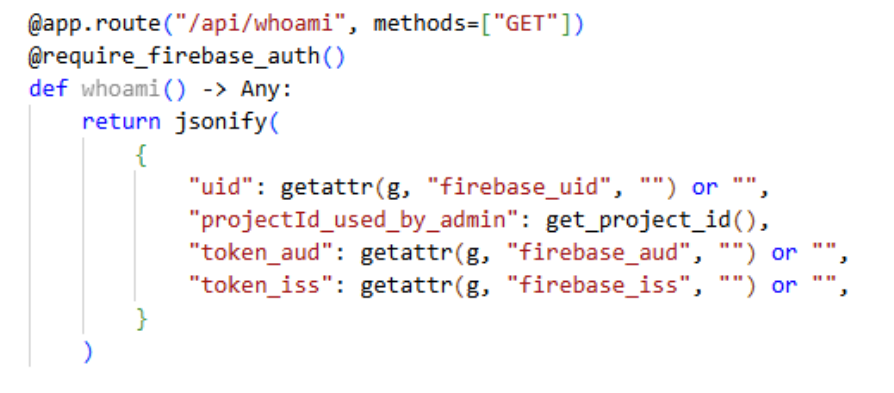}
\captionsetup{font=footnotesize,skip=4pt}
\captionof{figure}{Identity-inspection endpoint (original Figure 5.1). }
\end{minipage}\hfill
\begin{minipage}[t]{.48\linewidth}
\vspace{0pt}\centering
\includegraphics[width=\linewidth,height=.24\textheight,keepaspectratio]{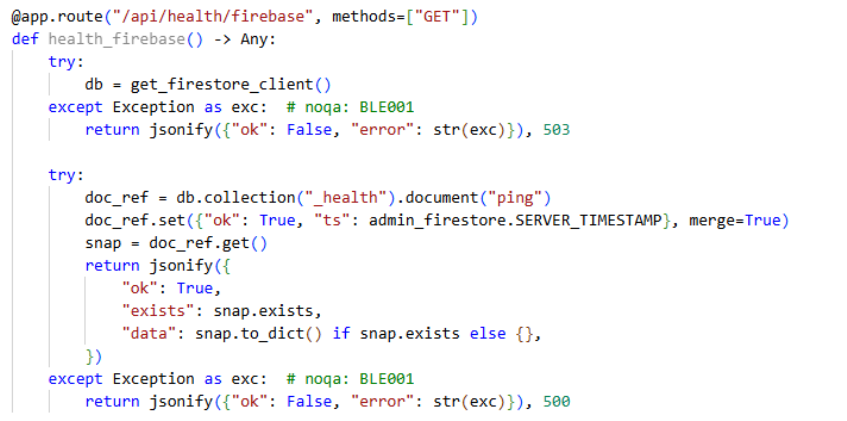}
\captionsetup{font=footnotesize,skip=4pt}
\captionof{figure}{Firebase health-check endpoint (original Figure 5.2). }
\end{minipage}\par\vspace{9pt}
\par
\noindent
\begin{minipage}[t]{.94\linewidth}
\vspace{0pt}\centering
\includegraphics[width=\linewidth,height=.46\textheight,keepaspectratio]{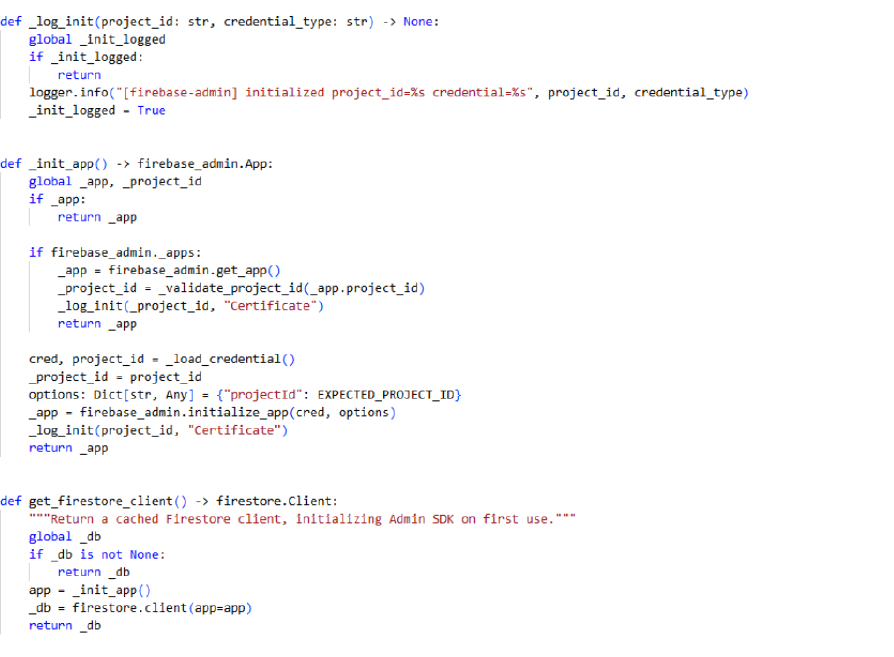}
\captionsetup{font=footnotesize,skip=4pt}
\captionof{figure}{Firebase initialization excerpt (original Figure 5.3). }
\end{minipage}\par\vspace{9pt}
\clearpage
\subsection*{Questionnaire response and session handling}
\par
\noindent
\begin{minipage}[t]{.52\linewidth}
\vspace{0pt}\centering
\includegraphics[width=\linewidth,height=.76\textheight,keepaspectratio]{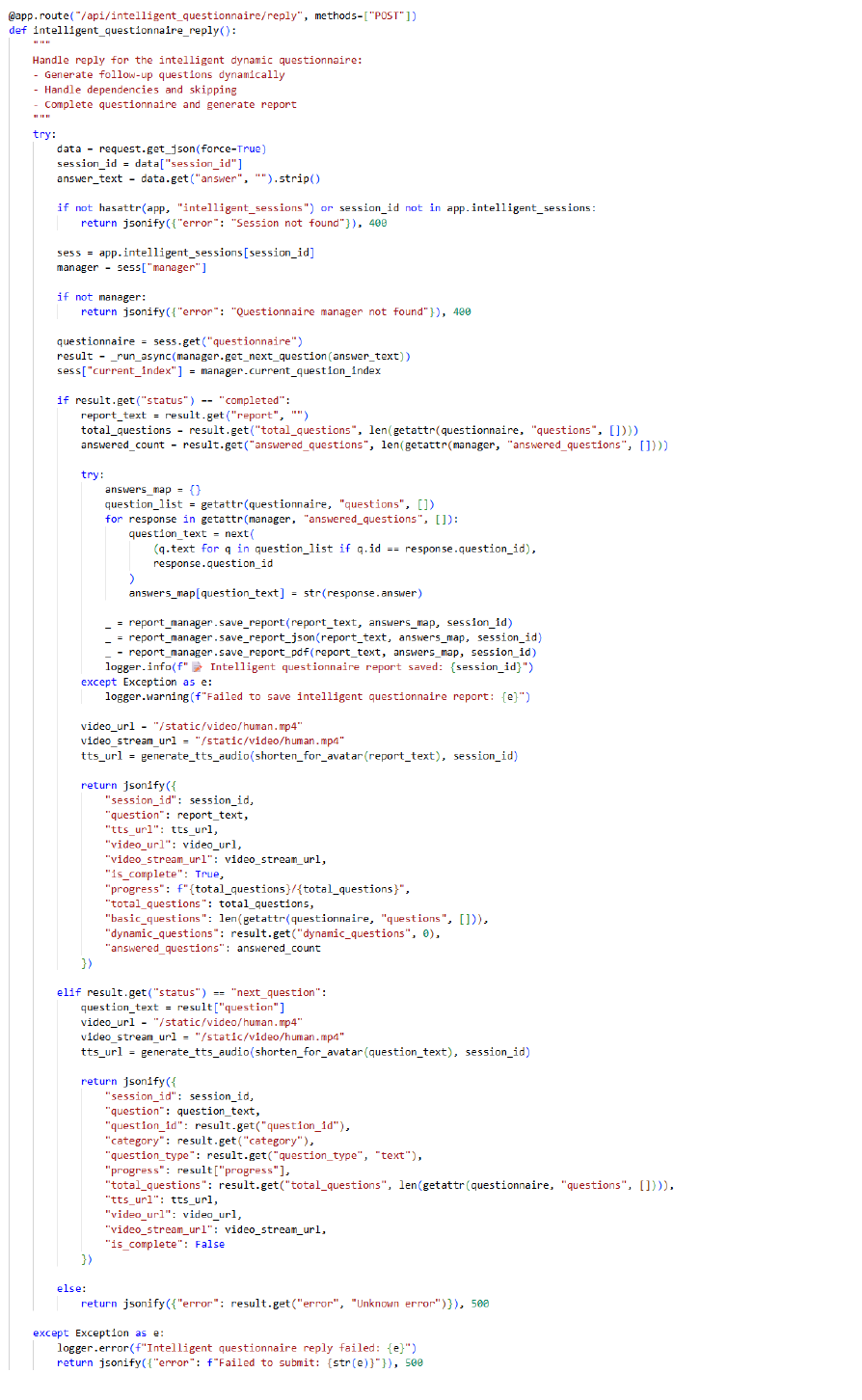}
\captionsetup{font=footnotesize,skip=4pt}
\captionof{figure}{Questionnaire reply-handler excerpt (original Figure 5.4). This image is a partial code excerpt; dependency and runtime context are not supplied.}
\end{minipage}\hfill
\begin{minipage}[t]{.44\linewidth}\vspace{0pt}
\begin{minipage}[t]{1\linewidth}
\vspace{0pt}\centering
\includegraphics[width=\linewidth,height=.25\textheight,keepaspectratio]{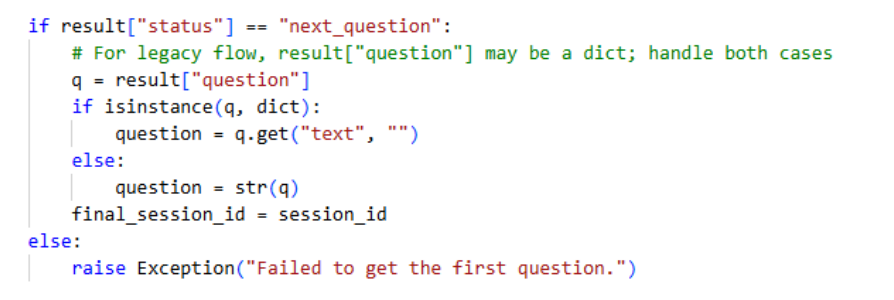}
\captionsetup{font=footnotesize,skip=4pt}
\captionof{figure}{Next-question response parsing (original Figure 5.5). }
\end{minipage}\par\vspace{14pt}
\begin{minipage}[t]{1\linewidth}
\vspace{0pt}\centering
\includegraphics[width=\linewidth,height=.30\textheight,keepaspectratio]{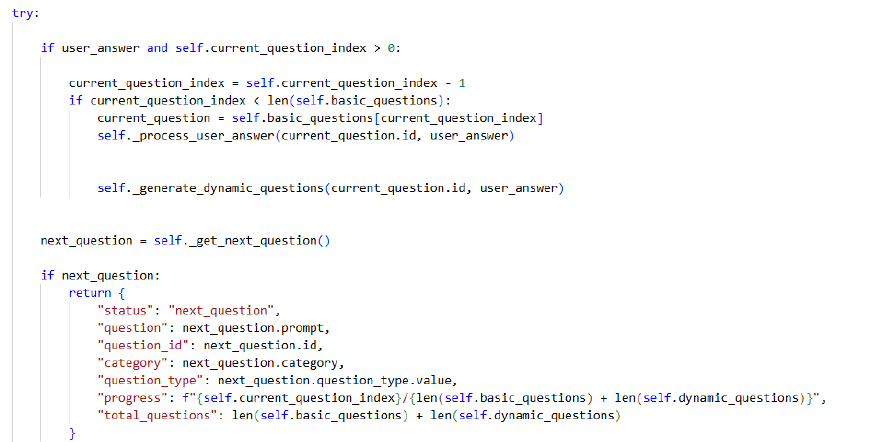}
\captionsetup{font=footnotesize,skip=4pt}
\captionof{figure}{Dynamic question and state-update excerpt (original Figure 5.6). }
\end{minipage}\end{minipage}\par\vspace{9pt}
\clearpage
\subsection*{Validation, risk aggregation, and model request}
\par
\noindent
\begin{minipage}[t]{.96\linewidth}
\vspace{0pt}\centering
\includegraphics[width=\linewidth,height=.20\textheight,keepaspectratio]{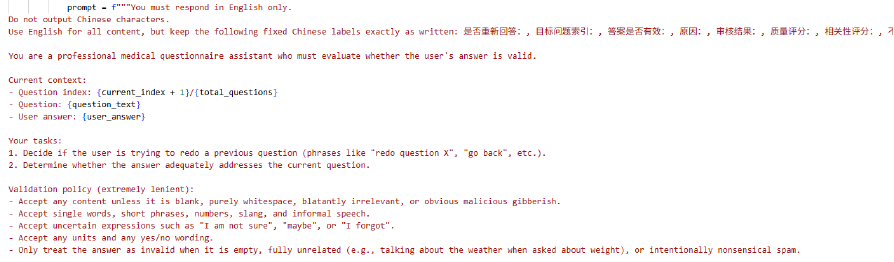}
\captionsetup{font=footnotesize,skip=4pt}
\captionof{figure}{Answer-validation prompt excerpt (original Figure 5.7). }
\end{minipage}\par\vspace{9pt}
\par
\noindent
\begin{minipage}[t]{.96\linewidth}
\vspace{0pt}\centering
\includegraphics[width=\linewidth,height=.37\textheight,keepaspectratio]{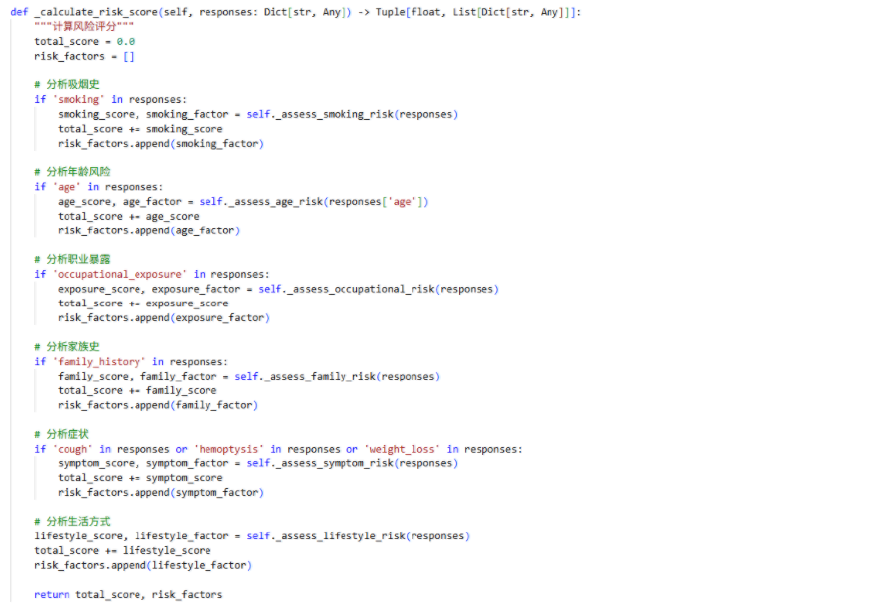}
\captionsetup{font=footnotesize,skip=4pt}
\captionof{figure}{Risk-factor aggregation excerpt (original Figure 5.8). Factor aggregation is visible, but the helper functions and category thresholds are not completely specified.}
\end{minipage}\par\vspace{9pt}
\par
\noindent
\begin{minipage}[t]{.96\linewidth}
\vspace{0pt}\centering
\includegraphics[width=\linewidth,height=.20\textheight,keepaspectratio]{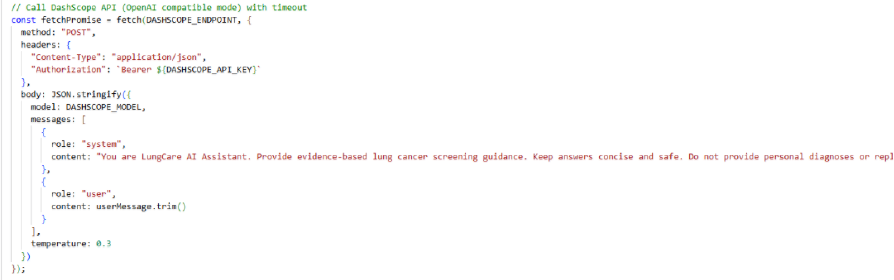}
\captionsetup{font=footnotesize,skip=4pt}
\captionof{figure}{Language-model request excerpt (original Figure 5.9). The screenshot illustrates a model request; it does not independently document retrieval quality or a complete retrieval pipeline.}
\end{minipage}\par\vspace{9pt}
\clearpage
\section{Requirement-Elicitation Instruments}
These materials concern requirements gathering. They are not the clinical questionnaire and do not report participant responses or empirical findings. The English survey is cropped to omit the signed-in account toolbar.
\par
\noindent
\begin{minipage}[t]{.54\linewidth}
\vspace{0pt}\centering
\includegraphics[width=\linewidth,height=.72\textheight,keepaspectratio]{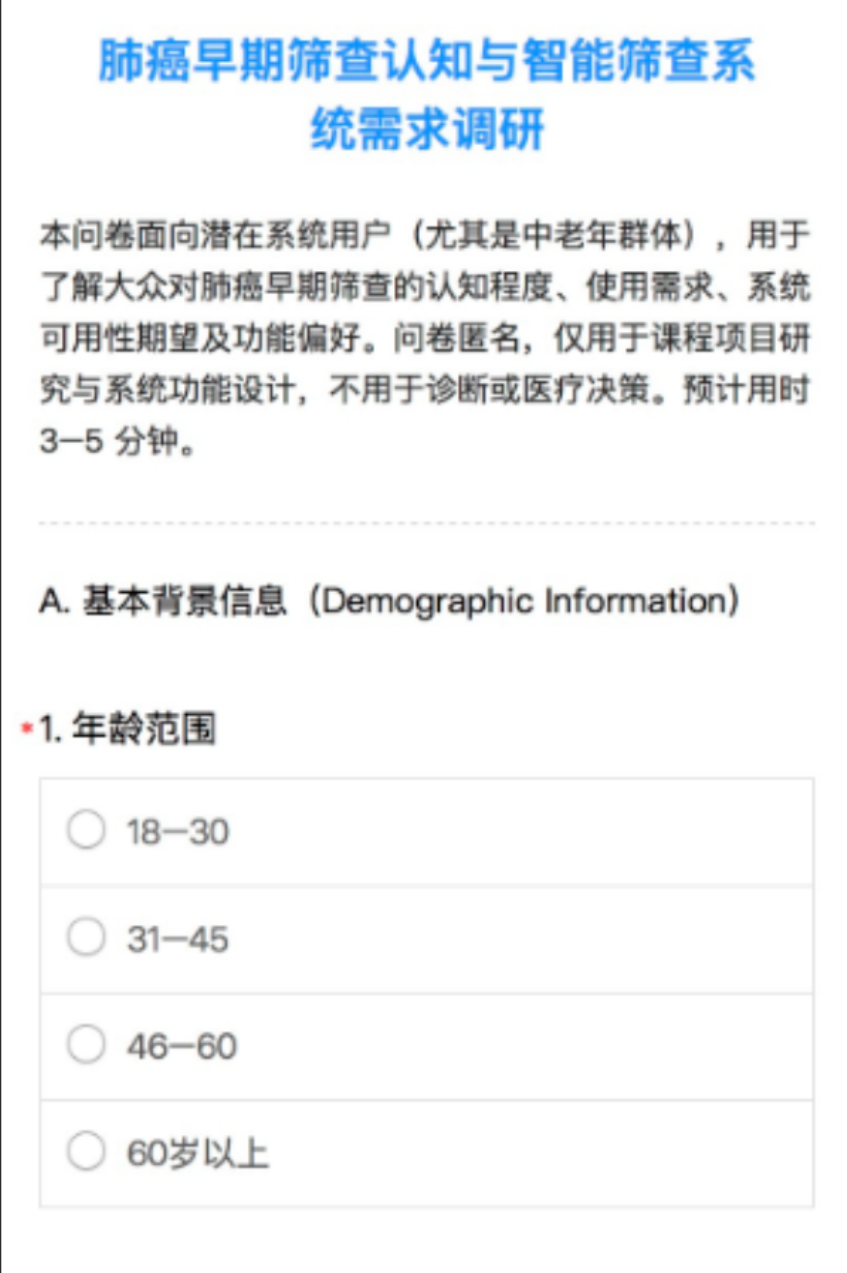}
\captionsetup{font=footnotesize,skip=4pt}
\captionof{figure}{Chinese requirement-survey screenshot (original Figure A1). This is a requirement-elicitation instrument, not the clinical risk questionnaire and not a set of survey results.}
\end{minipage}\hfill
\begin{minipage}[t]{.42\linewidth}
\vspace{0pt}\centering
\includegraphics[width=\linewidth,height=.30\textheight,keepaspectratio]{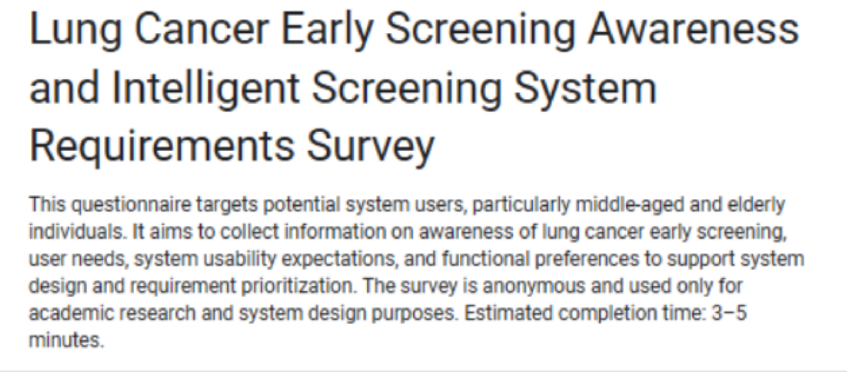}
\captionsetup{font=footnotesize,skip=4pt}
\captionof{figure}{English requirement-survey screenshot (original Figure A2). Only the survey description is reproduced; the signed-in account toolbar and lower navigation are omitted. No survey responses or findings are shown.}
\end{minipage}\par\vspace{9pt}
\clearpage
\subsection*{Semi-structured interview guide}
The source supplies the following interview topics but no participant inventory, transcripts, analysis, or findings. They are reproduced as a design instrument rather than evidence of a completed empirical study.
\begin{enumerate}[itemsep=5pt]
\item What is the participant's role or background, such as caregiving, medical study, healthcare, or health management?
\item Has the participant been involved in lung disease screening, diagnosis, or follow-up?
\item How is lung cancer screening conducted in the participant's experience?
\item Which screening or follow-up steps are most likely to be delayed or overlooked?
\item What difficulties do middle-aged or older people encounter during screening?
\item Could an online questionnaire support an initial screening-related interaction, and why?
\item What should a risk-information report include, such as its category, explanation, and next-step information?
\item Should AI serve as decision support, education and guidance, or neither?
\item How should the interface explain that generated output does not replace medical diagnosis?
\item Which decisions must remain the responsibility of medical professionals?
\item Should doctors be able to inspect historical reports and trends, and why?
\item What would require improvement or careful management before practical use?
\item What additional concerns or recommendations does the participant have?
\end{enumerate}